\documentclass{article}

\PassOptionsToPackage{numbers}{natbib}
\usepackage[main, final]{neurips_2026}

\usepackage[utf8]{inputenc} %
\usepackage[T1]{fontenc}    %
\usepackage{hyperref}       %
\usepackage{url}            %
\usepackage{booktabs}       %
\usepackage{amsfonts}       %
\usepackage{nicefrac}       %
\usepackage{microtype}      %
\usepackage{xcolor}         %

\usepackage{enumitem}

\usepackage{multirow}
\usepackage{longtable}
\usepackage{makecell}
\usepackage{caption}
\usepackage{tikz}
\usepackage{colortbl}
\usepackage{arydshln}
\usepackage{pifont}
\usepackage{array}
\usepackage{float}
\usepackage{wrapfig}
\usepackage{tabularx}

\usepackage{subcaption}
\usepackage[most]{tcolorbox}
\usepackage{listings}
\usepackage[capitalize,nameinlink]{cleveref}

\lstdefinestyle{promptstyle}{
  basicstyle=\ttfamily\small,
  breaklines=true,        %
  breakatwhitespace=false,
  columns=fullflexible,   %
  keepspaces=true,
}

\definecolor{purpletext}{RGB}{125, 60, 152}
\definecolor{neongreen}{RGB}{0, 155, 127}
\newcommand{\TextCircle}[1][0.7]{%
    \tikz[baseline=(char.base)]\node[shape=circle,draw=black,inner sep=1.5pt,line width=0.5pt,fill=purpletext,text=white,scale=#1] (char) {T};\hspace{-1pt}
}
\newcommand{\ImageCircle}[1][0.76]{%
    \tikz[baseline=(char.base)]\node[shape=circle,draw=black,inner sep=1.5pt,line width=0.5pt,fill=neongreen,text=white,scale=#1] (char) {I};\hspace{-1pt}
}

\newcommand\mypara[1]{\noindent\textbf{#1}}

\title{\emph{Reading, Not Thinking:} \\Understanding and Bridging the Modality Gap When Text Becomes Pixels in Multimodal LLMs}

\author{%
  Kaiser Sun\thanks{Equal contribution} \\
  Johns Hopkins University \\
  \texttt{hsun74@cs.jhu.edu} \\
  \And
  Xiaochuang Yuan\footnotemark[1] \\
  Amazon \\
  \And
  Hongjun Liu\footnotemark[1] \\
  New York University \\
  \AND
  Chen Zhao \\
  New York University \\
  \And
  Cheng Zhang \\
  Texas A\&M University \\
  \AND
  Mark Dredze \\
  Johns Hopkins University \\
  \And
  Fan Bai \\
  Johns Hopkins University \\
}

\begin{document}

\raggedbottom

\maketitle

\vspace{-5mm}
\begin{abstract}
    Multimodal large language models (MLLMs) can process text presented as images, yet they often perform worse than when the same content is provided as textual tokens.
    We systematically diagnose this ``modality gap'' by evaluating seven MLLMs across seven benchmarks in five input modes, spanning both synthetically rendered text and realistic document images from arXiv PDFs to Wikipedia pages.
    We find that the gap is highly sensitive to rendering choices such as font and resolution, and that natural document images often exhibit much smaller gaps, suggesting the performance difference partly reflects evaluation artifacts rather than fundamental limitations. 
    Through a grounded-theory error analysis of over 5,000 examples, we identify the primary cause: image input alone suppresses reasoning effort, with most models producing 8.6--85$\times$ shorter outputs that skip step-by-step computation or reasoning.
    The reluctance to reason, not a failure of perception or knowledge retrieval, drives the performance gap, particularly on tasks requiring multi-step reasoning.
    We show that a simple lightweight self-distillation method by fine-tuning models on their own text-mode reasoning traces paired with image inputs closes this gap, raising image-mode accuracy to match or exceed text-mode performance with over 50\% improvement, and the gains transfer to unseen benchmarks without catastrophic forgetting.
    Overall, our results and analyses provide a systematic understanding of the modality gap and suggest a practical path toward improving visual text understanding in multimodal language models.
    Code and data: \url{https://github.com/KaiserWhoLearns/ReadingNotThinking}.
\end{abstract}

\vspace{-6mm}
\begin{figure}[h!]
\centering
\includegraphics[width=0.83\linewidth]{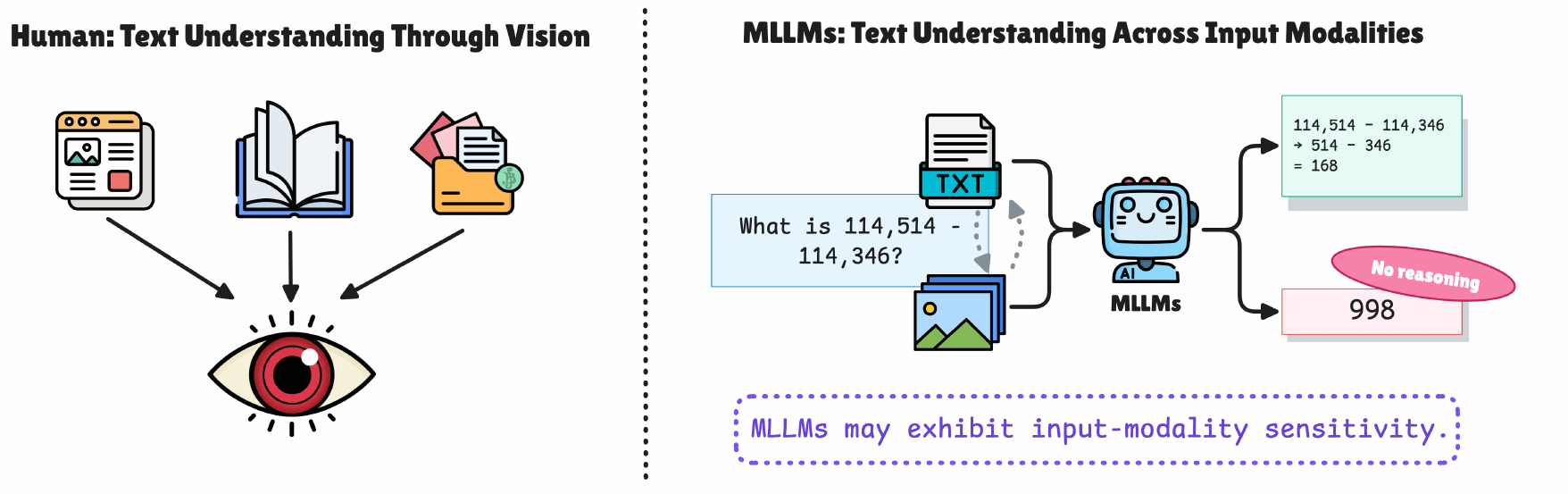}
\caption{Humans read text visually across diverse sources (e.g., books, webpages, documents), whereas MLLMs may produce different predictions for identical content presented in different formats.}
\label{fig:teaser}
\vspace{-10pt}
\end{figure}

\section{Introduction}
\label{sec:intro}

Humans read text through vision: we perceive words as visual patterns on pages, screens, and signs, not as sequences of abstract symbols~\cite{dehaene2007cultural, dehaene2018emergence}.
Modern multimodal large language models (MLLMs), however, process text and images through separate pathways, compressing visual input into tokens before fusing it with textual representations~\cite{liu2023llava, bai2023qwenvlversatilevisionlanguagemodel, chen2024internvl, agrawal2024pixtral, team2025gemma}.
When the same textual content is presented as an image rather than as text tokens, these models often produce different, and frequently worse, predictions (\Cref{fig:teaser}).
This ``modality gap'' has been documented across math, coding, and knowledge tasks~\citep{lyu2025pixelworldfarperceivingpixels, vansprang2025crossmodal}, and motivates a growing line of work on visual text processing, including screenshot language models~\citep{rust2023language, gao2024improvinglanguageunderstandingscreenshots}, vision-centric tokenization~\citep{xing2025seetok, cheng2025glyph}, and unified multimodal architectures~\citep{wang2026emu3}.

Yet existing work predominantly focuses on either \emph{documenting} the gap or \emph{building new systems} to circumvent it, without systematically explaining \emph{why} the gap arises in current MLLMs, characterizing \emph{when} visual input helps versus hinders, or showing \emph{how} the gap can be closed with minimal intervention.
A further limitation is that existing evaluations rely almost exclusively on \emph{synthetically rendered} text images, potentially confounding rendering artifacts with genuine perceptual deficits.

\begin{figure*}[!t]
    \centering
    \includegraphics[width=0.85\linewidth]{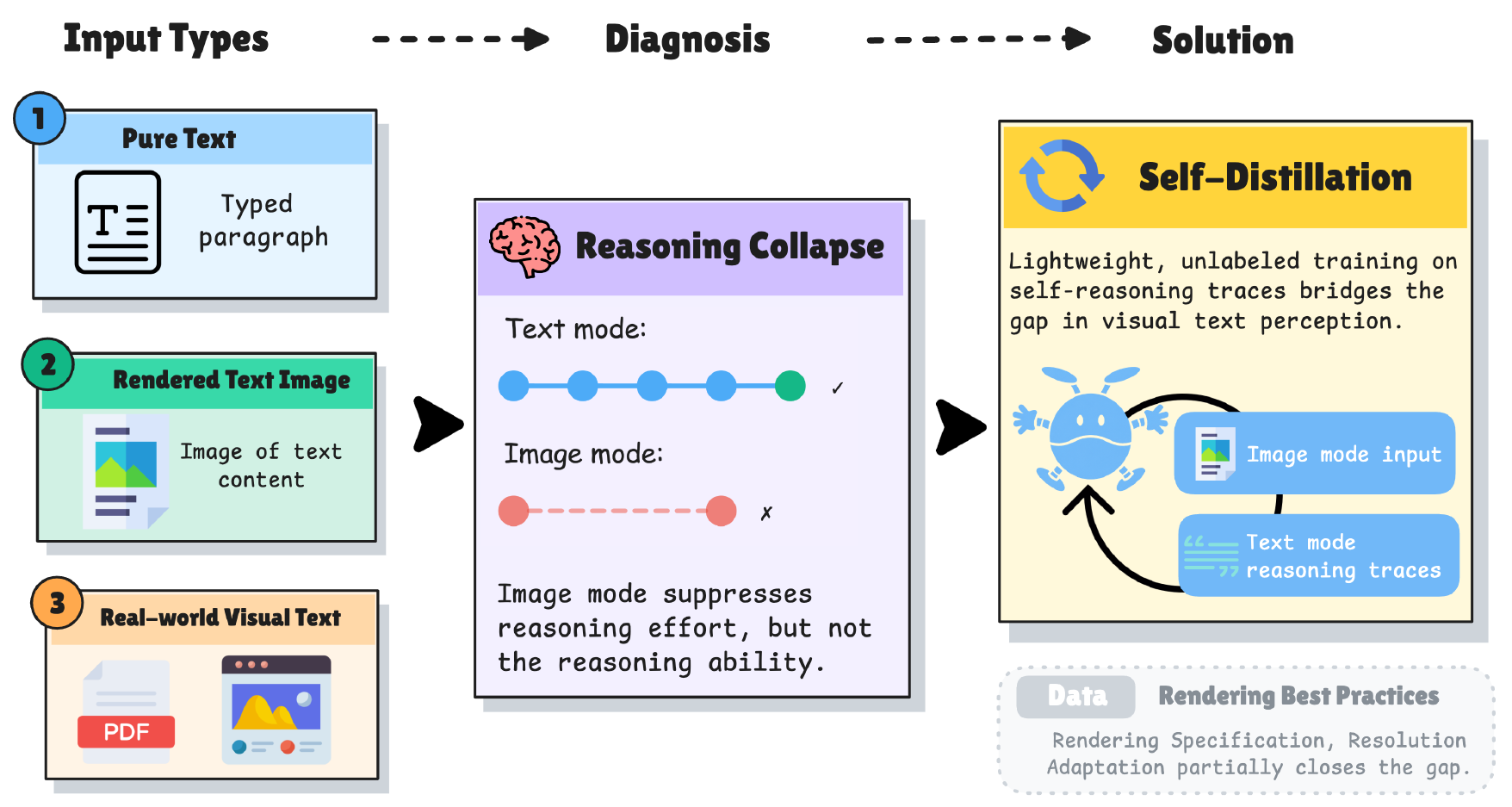}
    \caption{
    Diagnosing the modality gap in visual text understanding.
    We evaluate MLLMs across five input modes, including pure text, rendered text images, real-world visual text, and two OCR-based diagnostic settings (\texttt{OCR-1P} and \texttt{OCR-2P}). 
    Our error taxonomy reveals that while image modality amplifies reading and calculation errors, it leaves underlying reasoning capabilities largely intact but also reduces the likelihood of triggering a reasoning chain.
    We further propose remedies to bridge the performance gap, including rendering specification control, resolution-aware preprocessing, and LM-only self-distillation.
    }
    \label{fig:data-llu}
\end{figure*}

In this work, we diagnose the modality gap and show it can be substantially narrowed with a simple training recipe.
We evaluate seven MLLMs across seven benchmarks in five input modes (\Cref{fig:data-llu}), spanning synthetically rendered text and realistic document images from arXiv PDFs and Wikipedia pages.
Two OCR-based modes decompose the visual pipeline into reading and reasoning stages, isolating their respective contributions to the gap.

We first establish that the gap is task- and data-dependent.
On synthetic benchmarks, math tasks show gaps exceeding 60 points while knowledge-intensive tasks show modest drops.
On natural document images, MLLMs show much smaller performance differences and sometimes even exceed their text-mode performance.
Rendering choices such as font and resolution are critical confounds, with font alone swinging accuracy by up to 38 percentage points.
These findings suggest the gap partly reflects distributional mismatch rather than fundamental perceptual limitations.

To understand \emph{why} image mode degrades performance, we conduct a grounded-theory error analysis of over 5{,}000 errors.
We find that image mode triggers a reasoning collapse: models produce substantially shorter outputs, skipping step-by-step computation entirely.
The reluctance to reason in image mode, rather than a failure of visual perception or knowledge retrieval, is the major cause of the performance gap.
The error taxonomy confirms this: calculation and format errors increase under image input, while conceptual and reasoning errors remain unchanged.

Motivated by this diagnosis, we propose a self-distillation approach: training the model on its own text-mode reasoning traces paired with image inputs.
This simple approach, requiring only lightweight fine-tuning without architectural changes, raises image-mode accuracy to match or exceed text-mode performance.
We validate self-distillation on two open-source architectures, and show the gains transfer to unseen benchmarks without catastrophic forgetting.
Overall, our contributions are:
\begin{itemize}[leftmargin=*, itemsep=2pt]
\item We provide a comprehensive evaluation across seven models, seven benchmarks, and five input modes, showing that the gap is task-dependent and that rendering choices are critical confounds.
\item We identify reasoning collapse as the root cause of the modality gap: models are far less likely to produce step-by-step reasoning when the input is visual, leading to cascading errors.
\item We propose a lightweight but effective solution that closes the gap by training on text-mode reasoning traces paired with image inputs, without requiring architectural changes or external data.
\end{itemize}

\section{Related Work}
\label{sec:related_work}

\mypara{Visual Text Processing.}
A growing body of work processes language through visual representations.
Screenshot language models~\citep{rust2023language, gao2024improvinglanguageunderstandingscreenshots} train on rendered text with masked patch prediction, while vision-centric tokenization~\citep{xing2025seetok, cheng2025glyph} compresses context by 3--4$\times$ via rendering.
Unified architectures such as Emu3~\citep{wang2026emu3} perform next-token prediction over a shared image-text space.
These works focus on \emph{building better systems}; in contrast, we diagnose \emph{why} existing MLLMs fail on visual text and show the gap can be narrowed with minimal training.

\mypara{Evaluating MLLMs on Visual Text.}
Unlike scene-text understanding~\citep{singh2019vqamodelsread,biten2019scenetextvisualquestion,sidorov2020textcapsdatasetimagecaptioning}, which tests recognition in natural images, recent work evaluates \emph{reasoning} over visually presented text.
PixelWorld~\citep{lyu2025pixelworldfarperceivingpixels} and VISTA-Bench~\citep{liu2026vistabench} document task-dependent gaps, while Li et al.~\citep{li-etal-2025-text} show that rendering long contexts as images can halve token counts.
Concurrent work, Sprang et al.~\citep{vansprang2025crossmodal}, measure cross-modal consistency but do not explain \emph{why} inconsistency arises or how to reduce it.
We go beyond measurement: we evaluate on realistic documents (Wikipedia screenshots, arXiv PDFs), point out the anomaly within existing evaluations, provide an error taxonomy identifying reasoning collapse as the root cause, and propose self-distillation to close the gap.

\mypara{Closing the Modality Gap.}
While the works above establish that a text-versus-image gap exists, relatively little work addresses \emph{how to close} the gap in existing MLLMs.
Wang et al.~\citep{wang2025curepoison} show architecture-dependent effects of embedding text in images but propose no remedy; Hu et al.~\citep{hu2024arcvision} achieve strong ARC performance with visual priors alone but require training from scratch.
Our self-distillation trains the model on its own text-mode reasoning traces paired with image inputs.
This connects to cross-modal distillation~\citep{gupta2015crossmodaldistillationsupervision, tian2022contrastiverepresentationdistillation}, but uniquely uses the \emph{same} model as both teacher (text mode) and student (image mode).

\section{Evaluation Setup}
\label{sec:setup}

\subsection{Datasets}

As discussed in Section~\ref{sec:related_work}, our focus is on \emph{reasoning} over visually presented text rather than scene-text recognition. We include the following datasets:

\mypara{Synthetic-image datasets.}
Five benchmarks from prior visual text evaluations exhibit clear modality gaps, allowing us to examine their source.
\textbf{MMLU}~\cite{hendryckstest2021,hendrycks2021ethics} contains multiple-choice questions across 57 knowledge domains.
\textbf{ARC}~\cite{allenai:arc} includes challenging grade-school science questions.
\textbf{GPQA}~\cite{rein2023gpqa} (main set) contains 448 graduate-level science questions that experts answer correctly but non-experts often fail.
\textbf{GSM8K}~\cite{cobbe2021gsm8k} consists of grade-school math word problems.
\textbf{HumanEval}~\cite{chen2021evaluating} evaluates Python code generation from natural-language descriptions.
Because these tasks are purely textual, their image counterparts are synthetically rendered.

\mypara{Natural-image datasets.}
Synthetic renderings may not reflect the visual distribution of real-world text.
We therefore include two datasets whose image inputs come from natural sources.
\textbf{QASPER}~\cite{Dasigi2021ADO} is a long-form QA dataset over scientific articles that often requires long context reasoning.\footnote{We remove questions that depend on tables and figures for a fair comparison across modalities. Results on the full test set are in the appendix.}
We obtain the original PDF files from arXiv and convert relevant pages into PNG images using the \textit{fitz} library.\footnote{\url{https://pypi.org/project/fitz/}}
\textbf{SQuAD (v2)}~\cite{rajpurkar-etal-2018-know} is an extractive QA dataset built over Wikipedia.
For each example, we capture a screenshot of the corresponding Wikipedia page to create a natural-image version of the context.
Additional details on dataset processing and image construction are provided in the appendix.

\mypara{Evaluation metrics.}
We follow the standard metric for each dataset, using pass@k for HumanEval and accuracy for the rest.
For QASPER and SQuAD, zero-shot MLLM inference does not align well with span-based metrics, because MLLMs are not trained to reproduce annotation artifacts such as exact spans.
Following recent information-seeking benchmarks~\cite{wei2025browsecompsimplechallengingbenchmark, pham2025sealqaraisingbarreasoning}, we adopt an LLM-as-judge protocol, where we use GPT-5 (\textit{gpt-5-2025-08-07}) to compare the model prediction with the gold answer and report accuracy.
Results using span-based F1 are provided in the appendix.

\subsection{Input Modalities}
\label{sec:expsetup:data}

We compare five input modalities, with examples shown in the appendix.
Prior work has examined \texttt{Pure Text}, \texttt{Pure Image}, and \texttt{Instr.+Image} separately; we place them in a unified comparison and additionally introduce two OCR-based settings that help isolate whether the modality gap stems from failures in \emph{reading} text from pixels, failures in \emph{reasoning} over the extracted content, or both.

\mypara{\texttt{Pure Text.}}
The model receives only the original textual content (e.g., question, answer options, passages, or function signatures), preceded by a task instruction when necessary. This serves as our reference modality.

\mypara{\texttt{Pure Image.}}
The identical textual content from \texttt{Pure Text} is rendered into a 1280$\times$720 image (white canvas, black text) following prior work~\cite{lyu2025pixelworldfarperceivingpixels, clavie2025readbenchmeasuringdensetext}; the model receives only this image.
For natural-image datasets, the task instruction is rendered and concatenated at the top of the document image.
Section~\ref{sec:results} examines how rendering parameters (font, resolution, compression) affect performance.

\mypara{\texttt{Instr.+Image.}}
The model receives the example content as an image and the task instruction as text.
This setting closely mirrors traditional VQA-style inputs, where language instructions are provided textually while the model must read text embedded in the image.

\mypara{\texttt{OCR-1P (one pass).}}
The model receives the same image as in \texttt{Pure Image} (with the task instruction rendered in it), plus a text prompt that asks it to first extract all text from the image and then solve the task, all within a single inference pass.
This tests whether explicitly prompting for text extraction improves downstream performance while still coupling reading and reasoning.

\mypara{\texttt{OCR-2P (two passes).}}
In this two-stage pipeline, the model first receives the image and is prompted \emph{only} to extract the textual content.
The extracted text is then fed back into the model as a new \texttt{Pure Text} query to solve the task.
By fully separating reading from reasoning, this setting isolates whether errors originate from the vision-based text extraction or from downstream task execution.

\subsection{Models}
\label{sec:setup:models}

We evaluate seven MLLMs selected to span different architectures, model scales, and access types.
Our selection includes six open-source models: Qwen2.5-VL (7B and 32B)~\cite{bai2025qwen25vltechnicalreport}, Qwen3-VL-8B~\cite{yang2025qwen3}, InternVL3-8B~\cite{chen2024expanding}, InternVL3.5-8B~\cite{wang2025internvl3}, and Pixtral-12B~\cite{agrawal2024pixtral12b}, as well as the proprietary GPT-5.2 (\textit{gpt-5.2-2025-12-11}).
The open-source models share a common architecture in which a ViT-based visual encoder converts input images into visual tokens that are fed into an autoregressive LLM decoder.
All models receive identical inputs for each modality, and inference parameters are listed in the appendix.

\section{Data Characteristics and Rendering Confounds}
\label{sec:results}
\subsection{Modality Gap on Synthetic and Natural Images}
\begin{table*}[t]
\centering
\scriptsize
\resizebox{\textwidth}{!}{%
\begin{tabular}{@{}cl|ccccc|cl|ccccc@{}}
\toprule
\textbf{Dataset} & \textbf{Model} & \makecell{\scriptsize\textbf{Pure}\\\scriptsize\textbf{Text}\\\scriptsize(\TextCircle)} & \makecell{\scriptsize\textbf{Instr.+}\\\scriptsize\textbf{Image}\\\scriptsize(\TextCircle\ImageCircle)} & \makecell{\scriptsize\textbf{Pure}\\\scriptsize\textbf{Image}\\\scriptsize(\ImageCircle)} & \makecell{\scriptsize\textbf{OCR-1P}\\\scriptsize(\ImageCircle→\TextCircle)} & \makecell{\scriptsize\textbf{OCR-2P}\\\scriptsize(\ImageCircle→\TextCircle)} & \textbf{Dataset} & \textbf{Model} & \makecell{\scriptsize\textbf{Pure}\\\scriptsize\textbf{Text}\\\scriptsize(\TextCircle)} & \makecell{\scriptsize\textbf{Instr.+}\\\scriptsize\textbf{Image}\\\scriptsize(\TextCircle\ImageCircle)} & \makecell{\scriptsize\textbf{Pure}\\\scriptsize\textbf{Image}\\\scriptsize(\ImageCircle)} & \makecell{\scriptsize\textbf{OCR-1P}\\\scriptsize(\ImageCircle→\TextCircle)} & \makecell{\scriptsize\textbf{OCR-2P}\\\scriptsize(\ImageCircle→\TextCircle)} \\\\
\midrule
\multirow{7}{*}{\textbf{MMLU}} & \tiny{GPT-5.2} & 92.33 & \cellcolor{blue!6}{91.06}$^{\dagger}$ & \cellcolor{blue!6}{90.93}$^{\dagger}$ & \cellcolor{blue!10}{86.54}$^{\dagger}$ & \cellcolor{blue!10}{86.72}$^{\dagger}$ & \multirow{7}{*}{\textbf{GPQA}} & \tiny{GPT-5.2} & 81.25 & 81.25 & \cellcolor{blue!6}{80.13} & \cellcolor{blue!8}{77.90} & \cellcolor{blue!6}{80.58} \\

 & \tiny{InternVL3-8B} & 59.83 & \cellcolor{red!13}{69.09}$^{\dagger}$ & \cellcolor{blue!27}{33.93}$^{\dagger}$ & \cellcolor{red!5}{60.40} & \cellcolor{red!15}{71.86}$^{\dagger}$ &  & \tiny{InternVL3-8B} & 33.71 & \cellcolor{blue!6}{32.14} & \cellcolor{blue!6}{32.81} & \cellcolor{blue!8}{30.58} & \cellcolor{blue!5}{33.26} \\

 & \tiny{InternVL3.5-8B} & 82.97 & \cellcolor{blue!6}{82.25} & \cellcolor{blue!7}{80.74}$^{\dagger}$ & \cellcolor{blue!9}{77.65}$^{\dagger}$ & \cellcolor{blue!7}{80.75}$^{\dagger}$ &  & \tiny{InternVL3.5-8B} & 42.63 & \cellcolor{blue!7}{40.40} & \cellcolor{blue!12}{34.60}$^{\dagger}$ & \cellcolor{blue!7}{40.18} & \cellcolor{red!8}{46.21} \\

 & \tiny{Pixtral-12B} & 37.89 & \cellcolor{red!8}{41.60}$^{\dagger}$ & \cellcolor{red!9}{42.76}$^{\dagger}$ & \cellcolor{red!15}{49.54}$^{\dagger}$ & \cellcolor{red!10}{44.25}$^{\dagger}$ &  & \tiny{Pixtral-12B} & 31.25 & \cellcolor{blue!6}{30.36} & \cellcolor{blue!8}{27.23} & \cellcolor{blue!6}{29.69} & \cellcolor{blue!6}{30.36} \\

 & \tiny{Qwen2.5-7B-VL} & 67.80 & \cellcolor{red!5}{67.97} & \cellcolor{blue!5}{67.36} & \cellcolor{blue!7}{65.46}$^{\dagger}$ & \cellcolor{blue!6}{67.08} &  & \tiny{Qwen2.5-7B-VL} & 34.15 & 34.15 & \cellcolor{blue!8}{30.58} & \cellcolor{blue!9}{29.02} & \cellcolor{blue!6}{32.59} \\

 & \tiny{Qwen2.5-32B-VL} & 78.34 & \cellcolor{blue!7}{75.47}$^{\dagger}$ & \cellcolor{blue!9}{73.79}$^{\dagger}$ & \cellcolor{blue!6}{76.55}$^{\dagger}$ & \cellcolor{red!7}{80.38}$^{\dagger}$ &  & \tiny{Qwen2.5-32B-VL} & 42.41 & \cellcolor{blue!6}{41.52} & \cellcolor{red!6}{43.97} & \cellcolor{red!6}{43.08} & \cellcolor{red!6}{43.30} \\

 & \tiny{Qwen3-VL-8B} & 78.44 & \cellcolor{blue!10}{72.29}$^{\dagger}$ & \cellcolor{blue!11}{71.01}$^{\dagger}$ & \cellcolor{blue!10}{72.98}$^{\dagger}$ & \cellcolor{blue!9}{73.61}$^{\dagger}$ &  & \tiny{Qwen3-VL-8B} & 43.75 & \cellcolor{blue!9}{39.29} & \cellcolor{blue!12}{35.04}$^{\dagger}$ & \cellcolor{blue!10}{37.28} & \cellcolor{blue!11}{36.38} \\

\midrule

\multirow{7}{*}{\textbf{ARC}} & \tiny{GPT-5.2} & 96.25 & \cellcolor{blue!8}{92.24}$^{\dagger}$ & \cellcolor{blue!6}{95.48} & \cellcolor{red!6}{97.70} & \cellcolor{red!6}{97.18} & \multirow{7}{*}{\textbf{GSM8K}} & \tiny{GPT-5.2} & 95.83 & \cellcolor{blue!14}{85.14}$^{\dagger}$ & \cellcolor{red!6}{96.51} & \cellcolor{blue!20}{77.86}$^{\dagger}$ & \cellcolor{blue!28}{68.16}$^{\dagger}$ \\

 & \tiny{InternVL3-8B} & 92.24 & \cellcolor{blue!6}{91.30} & \cellcolor{blue!7}{90.27} & \cellcolor{blue!10}{86.26}$^{\dagger}$ & \cellcolor{blue!6}{90.53} &  & \tiny{InternVL3-8B} & 78.01 & \cellcolor{blue!29}{49.51}$^{\dagger}$ & \cellcolor{blue!35}{42.53}$^{\dagger}$ & \cellcolor{blue!8}{73.92} & \cellcolor{red!13}{87.87}$^{\dagger}$ \\

 & \tiny{InternVL3.5-8B} & 92.66 & \cellcolor{blue!5}{92.41} & 92.66 & \cellcolor{red!6}{93.26} & \cellcolor{red!6}{93.52} &  & \tiny{InternVL3.5-8B} & 95.30 & \cellcolor{blue!5}{95.22} & 95.30 & \cellcolor{red!5}{95.60} & \cellcolor{blue!6}{94.47} \\

 & \tiny{Pixtral-12B} & 84.64 & \cellcolor{blue!16}{71.25}$^{\dagger}$ & \cellcolor{blue!24}{61.77}$^{\dagger}$ & \cellcolor{blue!15}{72.10}$^{\dagger}$ & \cellcolor{blue!43}{39.33}$^{\dagger}$ &  & \tiny{Pixtral-12B} & 3.79 & \cellcolor{red!6}{4.93} & \cellcolor{red!9}{8.11}$^{\dagger}$ & \cellcolor{red!55}{75.89}$^{\dagger}$ & \cellcolor{red!48}{55.12}$^{\dagger}$ \\

 & \tiny{Qwen2.5-7B-VL} & 87.20 & \cellcolor{blue!6}{85.92} & \cellcolor{red!7}{89.42} & \cellcolor{red!5}{87.71} & \cellcolor{blue!6}{86.35} &  & \tiny{Qwen2.5-7B-VL} & 19.48 & \cellcolor{red!6}{20.24} & \cellcolor{red!9}{24.79}$^{\dagger}$ & \cellcolor{red!6}{20.09} & \cellcolor{red!6}{20.17} \\

 & \tiny{Qwen2.5-32B-VL} & 93.09 & \cellcolor{blue!6}{91.30} & \cellcolor{blue!8}{90.02}$^{\dagger}$ & \cellcolor{red!6}{94.62} & \cellcolor{red!6}{94.37} &  & \tiny{Qwen2.5-32B-VL} & 95.07 & \cellcolor{blue!55}{4.47}$^{\dagger}$ & \cellcolor{blue!5}{94.92} & \cellcolor{blue!29}{66.41}$^{\dagger}$ & \cellcolor{red!6}{96.13} \\

 & \tiny{Qwen3-VL-8B} & 91.72 & \cellcolor{blue!5}{91.38} & \cellcolor{blue!6}{91.04} & \cellcolor{red!6}{92.92} & \cellcolor{red!6}{92.75} &  & \tiny{Qwen3-VL-8B} & 93.56 & \cellcolor{blue!50}{39.58}$^{\dagger}$ & \cellcolor{blue!55}{30.71}$^{\dagger}$ & \cellcolor{blue!36}{56.94}$^{\dagger}$ & \cellcolor{red!6}{94.77} \\

\bottomrule

\end{tabular}%
}%
\caption{Performance of different input modalities on four synthetic-image datasets (MMLU, ARC, GPQA, GSM8K). Cell shading indicates the performance difference relative to \texttt{Pure Text}: \colorbox{red!30}{red} indicates the modality outperforms text, \colorbox{blue!30}{blue} indicates it underperforms text, and stronger shading reflects larger gaps. All HumanEval results throughout this paper are reported in \cref{app:humaneval} due to small dataset size and high pass@1 variance~\citep{chen2021evaluating}. $^{\dagger}$: significant difference from Pure Text (Holm-adjusted $p<0.05$; Appendix~\ref{app:significance}).}
\label{tab:overall_across_tasks}
\vspace{-3mm}
\end{table*}

\begin{table*}[t]
\centering
\tiny
\begin{tabular}{@{}llccccc@{}}
\toprule
\textbf{Dataset} & \textbf{Model} &
\makecell{\textbf{Pure Text}\\(\TextCircle)} & \makecell{\textbf{Instr.+Image}\\(\TextCircle\ImageCircle)} & \makecell{\textbf{Pure}\\\textbf{Image}\\(\ImageCircle)} & \makecell{\textbf{OCR-2P}\\(\ImageCircle→\TextCircle)}  & \makecell{\textbf{OCR-1P}\\(\ImageCircle)}  \\
\midrule

\multirow{5}{*}{\textbf{QASPER}} & GPT-5.2 & 51.92 & \cellcolor{red!24}{75.10} & \cellcolor{red!26}{77.25} & - & \cellcolor{red!24}{74.27} \\
 & Qwen2.5-7B-VL & 30.49 & \cellcolor{red!35}{66.32} & \cellcolor{red!33}{64.38} & - & \cellcolor{red!27}{57.05} \\
 & Qwen3-VL-8B & 53.27 & \cellcolor{red!24}{76.25} & \cellcolor{red!20}{70.75} & - & \cellcolor{red!8}{56.89} \\
 & InternVL3-8B & 44.02 & \cellcolor{red!26}{69.36} & \cellcolor{red!14}{54.43} & - & \cellcolor{red!15}{56.15} \\
 & InternVL3.5-8B & 49.38 & \cellcolor{red!12}{57.66} & \cellcolor{red!15}{61.07} & - & \cellcolor{red!10}{55.68} \\

\midrule

\multirow{6}{*}{\textbf{SQuAD v2}} & GPT-5.2 & 97.50 & \cellcolor{blue!15}{85.50} & \cellcolor{blue!8}{94.00} & \cellcolor{blue!15}{85.50} & \cellcolor{blue!15}{85.50} \\
 & Qwen2.5-7B-VL & 80.00 & \cellcolor{red!7}{82.50} & \cellcolor{red!8}{83.50} & \cellcolor{blue!8}{77.00} & \cellcolor{red!6}{81.50} \\
 & Qwen3-VL-8B & 95.00 & \cellcolor{blue!13}{85.50} & \cellcolor{blue!14}{84.50} & \cellcolor{blue!13}{85.50} & \cellcolor{blue!14}{84.00} \\
 & InternVL3-8B & 82.50 & \cellcolor{red!6}{84.00} & \cellcolor{blue!12}{73.50} & \cellcolor{blue!10}{76.50} & \cellcolor{blue!16}{69.50} \\
 & InternVL3.5-8B & 98.00 & \cellcolor{blue!17}{83.50} & \cellcolor{blue!17}{84.00} & \cellcolor{blue!15}{86.00} & \cellcolor{blue!20}{79.50} \\
 & Pixtral-12B & 91.50 & \cellcolor{blue!14}{81.00} & \cellcolor{blue!12}{83.50} & \cellcolor{blue!12}{82.50} & \cellcolor{blue!11}{84.50} \\

\bottomrule
\end{tabular}
\caption{Performance of different input modalities on two real-image datasets \textsc{Qasper} and SQuAD v2. Cell shading indicates the performance difference relative to \texttt{Pure Text}: \colorbox{red!30}{red} indicates the modality outperforms text, \colorbox{blue!30}{blue} indicates it underperforms text. Image-based inputs such as \texttt{Pure Image} and \texttt{Instr.+Image} outperform \texttt{Pure Text} on \textsc{Qasper}, indicating that MLLMs can benefit from naturally occurring visual text.\protect\footnotemark}
\label{tab:qasper_results}
\vspace{0pt}
\end{table*}
\footnotetext{OCR-2P is omitted for QASPER due to the computational cost of running OCR on extremely long documents. Pixtral-12B is excluded from QASPER due to insufficient context length.}

\mypara{Performance on synthetic images.}
\cref{tab:overall_across_tasks} reports results on four synthetic-image datasets, with cell shading indicating the performance difference relative to \texttt{Pure Text} (red = better than text, blue = worse).
On MMLU, \texttt{Pure Text} outperforms \texttt{Pure Image} for six of seven models.
GPQA shows the same direction for six of seven models (Appendix~\ref{app:significance}).
On ARC, models still perform better on text, but the gap narrows: Qwen2.5-VL-7B achieves $89.42\%$ under \texttt{Pure Image}, and InternVL3.5-8B ties its text baseline.
GSM8K reveals the largest modality gaps overall, with drops exceeding 60 points for several models.
HumanEval results are reported separately in \cref{app:humaneval} due to the high variance inherent in pass@1 estimation from only 164 instances~\citep{chen2021evaluating}.
\vspace{-1mm}

\mypara{Performance on natural images.}
\cref{tab:qasper_results} reports results on QASPER and SQuAD, where image inputs come from naturally occurring documents rather than synthetic renderings.
On QASPER, the trend reverses: almost all models experience a performance gain when switching from text to image.
These gains likely reflect the alignment between real PDF images and the document-heavy pretraining data of modern MLLMs.
On SQuAD, the behavior is more model-dependent: Qwen2.5-VL-7B benefits from images, whereas GPT-5.2 sees a much larger drop.
On SQuAD, rendering the original context under \texttt{Instr.+Image} changes accuracy by $+5.1$ percentage points on average across seven models relative to \texttt{Pure Text}, so an image input need not incur a performance penaltys.

\vspace{-1mm}
\mypara{Divergent robustness across models.}
As shown by the cell shading in Tab.~\ref{tab:overall_across_tasks} and Tab.~\ref{tab:qasper_results},
MLLMs exhibit sharply different responses to modality changes.
InternVL3.5-8B is the most modality-robust model, maintaining near-zero or positive differences on nearly all benchmarks.
Qwen2.5-VL-7B also shows strong robustness, with image often being its better-performing modality.
These discrepancies may stem from differences in vision encoder capability; we therefore use the OCR-1P and OCR-2P settings to measure recognition quality directly.
\vspace{-10pt}

\subsection{Image Recognition Cannot Fully Explain the Modality Gap}
\begin{wrapfigure}{r}{0.49\linewidth}
    \centering
    \vspace{-7mm}
    \includegraphics[width=\linewidth]{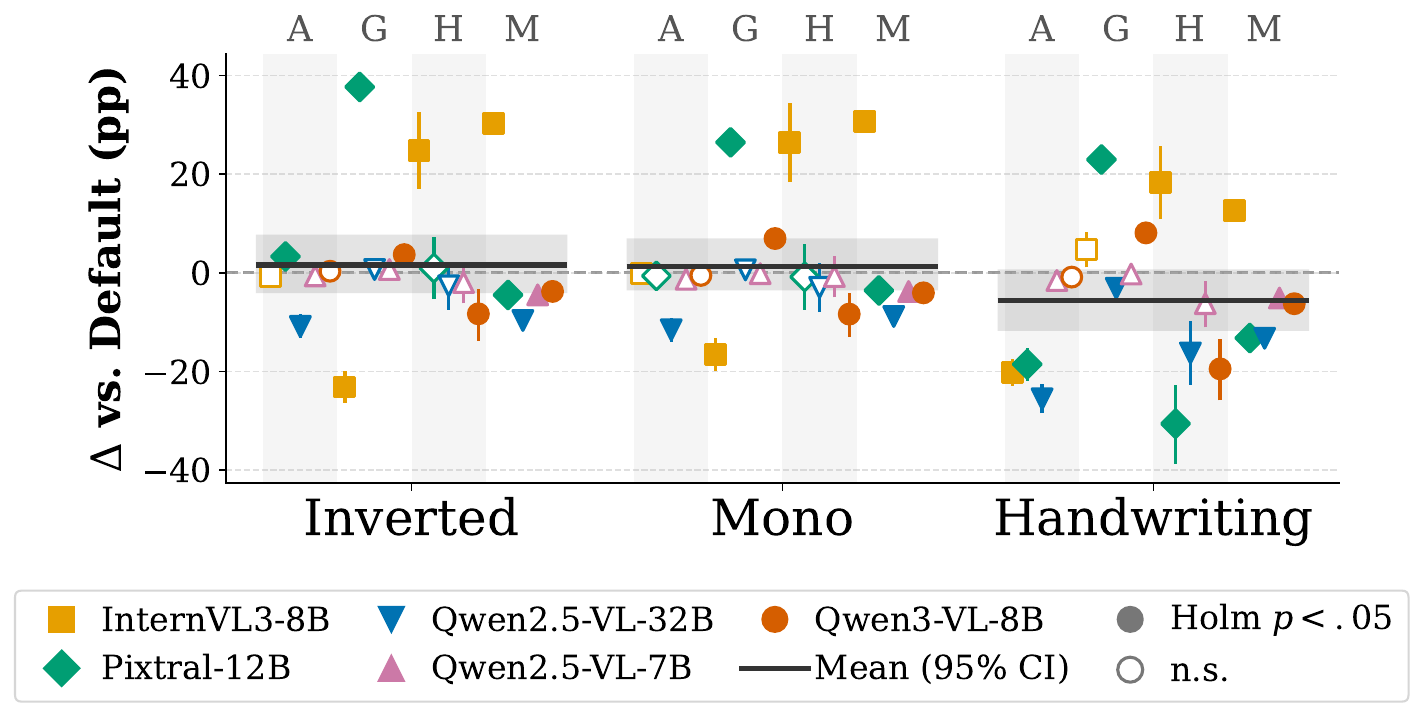}
    \caption{
    Accuracy change of each rendering relative to the default rendering. Each point is a dataset--model pair (A/G/H/M = ARC/GSM8K/HumanEval/MMLU) with its 95\% paired bootstrap CI; filled points are significant (Holm-corrected $p<0.05$).
    Handwriting is the only rendering with a negative mean change.
    }
    \label{fig:typefaceablation}
    \vspace{-15pt}
\end{wrapfigure}

The two OCR settings in \cref{tab:overall_across_tasks} reveal an asymmetric pattern across tasks.
\texttt{OCR-1P} matches or exceeds \texttt{Pure Image} in 16 of 28 cells, with the largest gains where \texttt{Pure Image} collapses (Pixtral-12B on GSM8K: 8.11 to 75.89; InternVL3-8B on MMLU: 33.93 to 60.40), although it hurts the strongest models on GSM8K (GPT-5.2: 96.51 to 77.86).
\texttt{OCR-2P} likewise recovers a portion of the performance: InternVL3-8B increases from 42.53 to 87.87 on GSM8K when switching from \texttt{Pure Image} to \texttt{OCR-2P}, and Pixtral-12B sees a 42-point improvement.
To quantify this relationship, we compute the correlation between \texttt{OCR-2P} word error rate and \texttt{OCR-2P} accuracy across all models and tasks, obtaining only $r = 0.163$.
This suggests that overall OCR quality is a poor predictor of image-mode performance; errors on a small number of task-critical tokens (e.g., operators, variable names) may matter more than aggregate recognition accuracy.

\subsection{Rendering Choices Confound Visual Evaluation}
\begin{wrapfigure}{r}{0.48\linewidth}
    \vspace{-15pt}
    \centering
    \includegraphics[width=\linewidth]{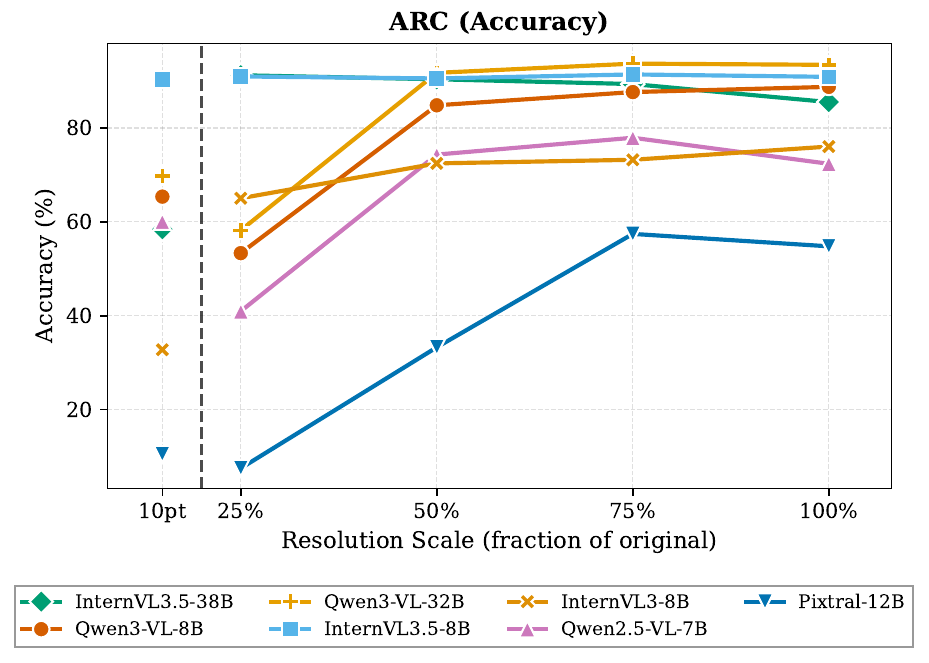}
    \caption{
    Performance vs.\ image resolution scale on ARC.
    The black vertical dashed line indicates the point where \texttt{Pure Image} consumes the same FLOPs as \texttt{Pure Text}.
    Other datasets are in \cref{app:resolution_curves}.
    }
    \label{fig:resolution_curves}
    \vspace{-5pt}
\end{wrapfigure}

\label{sec:rendering_confound}
The results above show that MLLMs struggle with synthetic renderings yet perform well on natural document images, suggesting a distributional mismatch between evaluation images and pretraining data.

\mypara{Font choice matters.}
We render text in four styles: default (NotoSansMath), inverted colors, monospaced (DejaVuSans Mono), and handwriting (Priestacy).
Figure~\ref{fig:typefaceablation} shows that font choice can swing accuracy by up to 38 percentage points, with handwriting causing the largest degradation.
Paired comparisons against the default rendering are significant in 38 of 60 cases after Holm correction across all 60 tests ($p_{\mathrm{adj}}<0.05$): 12/20 for inverted colors, 11/20 for monospaced text, and 15/20 for handwriting (Appendix~\ref{app:rendering_significance}).
This directly contradicts Sprang et al.~\citep{vansprang2025crossmodal}, who report no effect of font on cross-modal consistency.
The discrepancy likely reflects our focus on \emph{task accuracy} rather than response consistency: fonts common in pretraining (inverted, monospaced) cause minimal harm, while out-of-distribution fonts (handwriting) severely degrade performance.

To assess whether the handwriting degradation reflects difficulty reading the text, we separately test transcription on 50 instances per model--dataset pair for two models and three datasets (Appendix~\ref{app:handwriting_ocr}).
WER is only $0.9$--$6.1\%$ and CER $0.5$--$1.7\%$, with no significant correlation between per-item WER and image-mode answer correctness in any pair (all $p\geq 0.13$). 
These results suggest that recognition errors alone do not explain the handwriting degradation.

\mypara{Resolution sensitivity.}
Figure~\ref{fig:resolution_curves} shows that most models maintain accuracy at 0.50$\times$--1.0$\times$ resolution but degrade sharply below this threshold.
InternVL3.5, with its Visual Resolution Router~\cite{wang2025internvl3}, is uniquely resolution-invariant.
Notably, a compact 10pt rendering outperforms higher-resolution counterparts, suggesting that smart compression preserving critical features can mitigate resolution sensitivity.
Overall, font, color, and resolution are strong confounders that benchmarks should report alongside results.

\section{Error Analysis}
The aggregate performance numbers in Section~\ref{sec:results} show that image mode degrades accuracy, but not \emph{why}.
To move beyond surface-level accuracy comparisons, we conduct a systematic error analysis following \emph{grounded theory}~\citep{corbin2014basics}.
We classify over 5{,}000 individual errors across all models and six datasets under three input modes (\texttt{Pure Text}, \texttt{Pure Image}, \texttt{Instr.+Image}).

\subsection{Method}
Grounded theory~\citep{glaser1967discovery, corbin2014basics} is a widely adopted qualitative research methodology that has been a standard approach in sociology, healthcare, and education for over five decades.
Unlike hypothesis-driven methods, it derives theoretical frameworks inductively from data through iterative coding phases: \emph{open coding}, which segments raw observations into initial descriptive labels; \emph{constant comparison}, which refines codes by classifying new data against existing categories; and \emph{axial coding}, which groups related codes into higher-level categories.
Following Nelson's~\citep{nelson2020computational} computational grounded theory framework, which combines human interpretive skills with computational pattern recognition for rigorous and reproducible analysis, we adopt a human-in-the-loop variant.
Recent work has validated that LLM-assisted coding achieves high agreement with human experts when humans review model outputs~\cite{zhong2025hicode, wan2025noise, cemri2025multiagentllmsystemsfail, xiao2023supporting}.
In our implementation, GPT-5.2 performs the open coding and constant comparison while human annotators review \emph{every} proposed change and conduct the final axial coding to ensure coherence and validity.

\mypara{Phase 1: Open coding.}
We sample 1{,}000 errors uniformly across models, datasets, and modalities.
For each error, the model receives the original input (question and answer choices where applicable), ground-truth answer, and the model's raw response, and proposes a short failure description along with a candidate error label.
The model maintains a running set of codes and can either assign an instance to an existing code or propose a new one.
This phase produced 415 initial codes.
\vspace{-1mm}

\mypara{Phase 2: Constant comparison.}
New errors are randomly sampled and classified against existing codes.
For each instance, the model determines whether the error matches an existing code, requires a new code, or warrants updating an existing description.
A human annotator reviews every proposed new code, merge, or description update, approving or rejecting changes.
Saturation is reached when no new codes are added for 50 consecutive instances.
Codes occurring only once are pruned unless conceptually critical, and the corresponding errors are assigned to a miscellaneous category.
\vspace{-1mm}

\mypara{Phase 3: Axial coding.}
A human expert reviews all codes and groups them into organizing dimensions, producing the final taxonomy of primary error categories along with a miscellaneous category.
\vspace{-1mm}

\label{sec:error_analysis}
\begin{figure}[t]
    \centering
    \includegraphics[width=\linewidth]{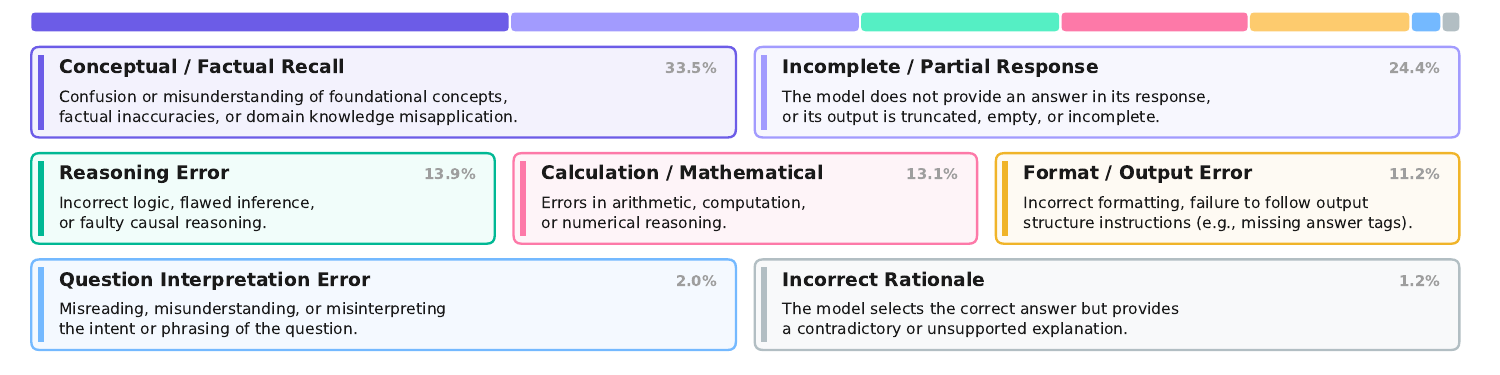}
    \caption{Error taxonomy derived from grounded-theory coding of 5{,}245 errors across all models and datasets. Percentages indicate overall frequency.}
    \label{fig:error_taxonomy}
\end{figure}

\subsection{Error Taxonomy and Findings}

The coding process above yields the taxonomy shown in Figure~\ref{fig:error_taxonomy}; an inter-annotator agreement study validates that this taxonomy can be reliably applied with moderate to substantial agreement ($\kappa = 0.57$--$0.69$, Appendix~\ref{app:human_annotation}).
We then classify up to 150 errors per model--dataset pair, yielding 5{,}245 errors across 40 pairs.

\mypara{Shortened reasoning under image input.}
The key finding is that models are far less likely to produce long-form reasoning in image mode.
On GSM8K, Qwen3-VL-8B averages 617 characters per response in text mode but only 32 in image mode; Pixtral averages 115 characters in text mode but only 22 in image mode.
Rather than working through problems step-by-step, models in image mode jump directly to answers or exit reasoning early, often resulting in incorrect answers (see Figure~\ref{fig:reasoning_collapse_example} in Appendix for an example).
This reluctance to reason explicitly has cascading consequences: without intermediate computation steps, arithmetic mistakes go undetected and uncorrected, and the model loses the self-verification that chain-of-thought reasoning provides.

\begin{figure*}[t]
    \centering
    \begin{subfigure}[t]{0.53\textwidth}
        \centering
        \includegraphics[width=\linewidth]{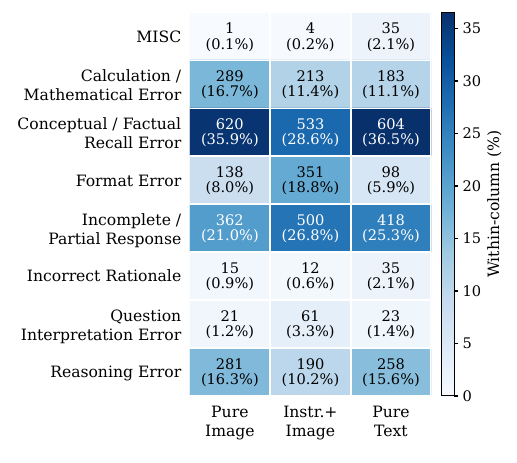}
        \caption{Error distribution by input mode.}
        \label{fig:error_by_mode}
    \end{subfigure}
    \hfill
    \begin{subfigure}[t]{0.44\textwidth}
        \centering
        \includegraphics[width=\linewidth]{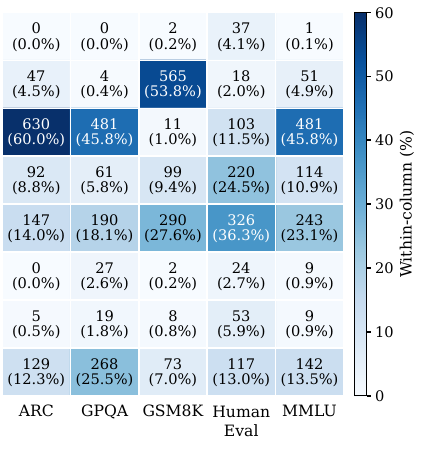}
        \caption{Error distribution by dataset.}
        \label{fig:error_by_dataset}
    \end{subfigure}
    \caption{Distribution of error categories across input modes (left) and datasets (right). Cell values show raw counts and within-column percentages.}
    \label{fig:error_analysis}
\end{figure*}

Comparing error distributions between image mode and text mode (Figure~\ref{fig:error_by_mode}), we observe a consistent pattern:
\emph{Calculation errors} make up a larger share of errors under image input (16.7\% vs.\ 11.1\%), and \emph{format errors} rise from 5.9\% to 8.0\%.
These are precisely the error types that reasoning helps prevent: step-by-step computation catches arithmetic mistakes, and longer outputs provide more opportunity to follow formatting instructions.
In contrast, \emph{conceptual/factual recall errors} and \emph{reasoning errors} remain virtually unchanged.
This asymmetry indicates that image mode does not impair the model's knowledge or logical capabilities; rather, it suppresses the reasoning process that would normally be applied.

\mypara{Dataset-specific error profiles.}
Output lengths before and after self-distillation are reported in Appendix~\ref{app:output_lengths}; benchmark-matched error comparisons are in Appendix~\ref{app:significance}.
The error distribution varies across tasks (Figure~\ref{fig:error_by_dataset}), reflecting their different requirements.
GSM8K errors are dominated by calculation failures, consistent with reasoning collapse having the largest impact on multi-step arithmetic.
ARC, GPQA, and MMLU show high rates of conceptual/factual recall errors, reflecting the knowledge-intensive nature of these benchmarks.
HumanEval exhibits a distinct profile with high rates of incomplete responses (36.3\%) and format errors (24.5\%), suggesting code generation is potentially sensitive to the output truncation that accompanies the reasoning collapse.

These findings establish that the modality gap is primarily driven by the reluctance to reason when only images are provided: image mode suppresses step-by-step reasoning, leading to cascading errors in calculation and formatting.
In Section~\ref{sec:self_distillation}, we show that training models to restore reasoning traces from image inputs substantially closes the gap.
\vspace{-2mm}

\section{Bridging the Modality Gap via Self-Distillation}
\label{sec:self_distillation}

Our error analysis (Section~\ref{sec:error_analysis}) identified reasoning collapse as the root cause of the modality gap: models produce dramatically shorter outputs in image mode, skipping step-by-step computation.
Our primary focus is training-time alignment with text-mode reasoning, which directly addresses the model's tendency to suppress reasoning under visual input.
For users with limited training resources, adding a chain-of-thought (CoT) prompt to the image offers a simple inference-time alternative, which we evaluate on a subset of model--benchmark pairs in Appendix~\ref{app:cot_prompting}.
To restore reasoning under visual input, we propose \emph{self-distillation}: training the model on its own text-mode reasoning traces paired with image inputs, teaching it to produce the same multi-step reasoning regardless of input modality.

We select GSM8K and MMLU because they exhibit large modality gaps and provide standard training splits.
For each model, we run inference in \texttt{Pure Text} mode to collect reasoning traces, retaining only those with correct final answers.
These traces serve as supervised targets, each paired with the corresponding \texttt{Pure Image} input.
Unlike standard distillation, which transfers soft labels, our setting requires transferring multi-step reasoning traces; ground-truth answers alone are insufficient since they lack the intermediate steps.
All experiments apply LoRA~\citep{hu2022lora} to the language model (freezing the vision encoder), with rank $r{=}64$, trained for 2 epochs with a learning rate of $2{\times}10^{-4}$ and an effective batch size of 16.

\begin{table*}[h]
\centering
\vspace{-2mm}
\scriptsize
\setlength{\tabcolsep}{1.5pt}
\begin{tabular}{@{}l l cc cc c|l l cc cc c@{}}
\toprule
& & \multicolumn{2}{c}{\textbf{Base}} & \multicolumn{2}{c}{\textbf{Distilled}} & & & & \multicolumn{2}{c}{\textbf{Base}} & \multicolumn{2}{c}{\textbf{Distilled}} & \\
\cmidrule(lr){3-4} \cmidrule(lr){5-6} \cmidrule(lr){10-11} \cmidrule(lr){12-13}
\textbf{Train} & \textbf{Eval} & \TextCircle & \ImageCircle & \TextCircle & \ImageCircle & \makecell{\scriptsize$|\Delta|$} & \textbf{Train} & \textbf{Eval} & \TextCircle & \ImageCircle & \TextCircle & \ImageCircle & \makecell{\scriptsize$|\Delta|$} \\
\midrule
\multicolumn{7}{@{}c|}{\textbf{Qwen3-VL-8B}} & \multicolumn{7}{c@{}}{\textbf{InternVL3-8B}} \\
\midrule
\multirow{4}{*}{GSM8K} & GSM8K & 93.56 & 30.71 & \textbf{94.09} & \textbf{92.72} & 62.85$^{\dagger}$${\to}$\textbf{1.37}$^{\ddagger}$ & \multirow{4}{*}{GSM8K} & GSM8K & 78.01 & 42.53 & \textbf{90.30} & \textbf{90.14} & 35.48$^{\dagger}$${\to}$\textbf{0.16}$^{\ddagger}$ \\
& ARC & 91.72 & 91.04 & 91.47 & 91.89 & 0.68${\to}$0.42 & & ARC & 92.24 & 90.27 & 91.38 & 89.33 & 1.97${\to}$2.05$^{\dagger}$ \\
& MMLU & 78.44 & 71.01 & 75.49 & 72.70 & 7.43$^{\dagger}$${\to}$2.79$^{\dagger}$$^{\ddagger}$ & & MMLU & 59.83 & 33.93 & \textbf{75.16} & 68.00 & 25.90$^{\dagger}$${\to}$7.16$^{\dagger}$$^{\ddagger}$ \\
& MMMU-Pro$^{\star}$ & 60.46 & 45.14 & 58.55 & 44.51 & -- & & MMMU-Pro$^{\star}$ & 48.09 & 33.35 & \textbf{54.51}$^{\dagger}$ & 32.54 & -- \\
\midrule
\noalign{\vskip 2pt}
\multirow{4}{*}{MMLU} & GSM8K & 93.56 & 30.71 & 93.63 & 85.06 & 62.85$^{\dagger}$${\to}$8.57$^{\dagger}$$^{\ddagger}$ & \multirow{4}{*}{MMLU} & GSM8K & 78.01 & 42.53 & 87.95 & 85.75 & 35.48$^{\dagger}$${\to}$2.20 \\
& ARC & 91.72 & 91.04 & 91.21 & 91.55 & 0.68${\to}$\textbf{0.34} & & ARC & 92.24 & 90.27 & 91.21 & \textbf{90.61} & 1.97${\to}$\textbf{0.60} \\
& MMLU & 78.44 & 71.01 & 72.20 & 71.73 & 7.43$^{\dagger}$${\to}$\textbf{0.47}$^{\ddagger}$ & & MMLU & 59.83 & 33.93 & 73.60 & \textbf{72.45} & 25.90$^{\dagger}$${\to}$\textbf{1.15}$^{\dagger}$$^{\ddagger}$ \\
& MMMU-Pro$^{\star}$ & 60.46 & 45.14 & 57.98 & 43.87 & -- & & MMMU-Pro$^{\star}$ & 48.09 & 33.35 & \textbf{55.26}$^{\dagger}$ & 34.68 & -- \\
\bottomrule
\end{tabular}
\vspace{2mm}
\caption{Self-distillation with Pure-Image inputs, Text-mode teachers, and LM-only LoRA. $|\Delta|$: absolute text--image gap (baseline$\to$distilled). $^{\star}$MMMU-Pro is a forgetting control: \TextCircle/\ImageCircle{} denote Standard/Vision, with no text--image gap. $^{\dagger}$ marks significant baseline/residual gaps or MMMU-Pro changes from baseline; $^{\ddagger}$ marks significant signed gap closure (Holm-adjusted $p<0.05$; Appendix~\ref{app:significance}).}
\label{tab:self_distill}
\end{table*}

\subsection{Results}
\vspace{-2mm}

\mypara{Self-distillation closes the modality gap.}
Table~\ref{tab:self_distill} presents results across two architectures (Qwen3-VL-8B and InternVL3-8B) and two training datasets (GSM8K and MMLU) with the highest modality gaps.
With a small amount of self-distillation, all four configurations substantially close the gap.
For both models and both training sets, self-distillation reduces the modality gap by an order of magnitude or more.
The largest improvement appears for Qwen3-VL-8B + GSM8K, where image-mode accuracy jumps from 30.71\% to 92.72\%, matching text-mode performance.
Text-mode performance is largely preserved, with one  exception (Qwen3-VL-8B trained on MMLU drops from 78.44\% to 72.20\% on MMLU text), indicating that self-distillation strengthens the visual-text alignment while largely preserving the original textual reasoning capability.
\vspace{-1mm}

\mypara{Transfer to unseen benchmarks.}
We evaluate each trained model on benchmarks not used during training to test whether self-distillation causes catastrophic forgetting.
Across all four configurations, models preserve or improve performance on held-out tasks, with the text-image gap narrowing on most benchmarks.
For instance, InternVL3-8B trained on GSM8K improves MMLU image-mode from 33.93\% to 68.00\%; Qwen3-VL-8B trained on GSM8K maintains HumanEval image-mode performance (\cref{app:humaneval}).
Our results above suggest that self-distillation teaches general visual-text alignment rather than task-specific shortcuts.

We additionally evaluate MMMU-Pro~\citep{yue2025mmmu} using its official \texttt{Standard} and \texttt{Vision} configurations with unmodified input formats, greedy decoding, a maximum of $8{,}192$ output tokens, strict matching of the extracted answer letter to the gold answer, and no additional instruction prompt.
Due to the nature of the task, MMMU-Pro has no \texttt{Pure Text} mode and serves only as a held-out catastrophic-forgetting control; our gap-closure claim rests on GSM8K, MMLU, ARC, and HumanEval, where identical content is available in both text and image form.
Across the four model--training-set configurations, MMMU-Pro \texttt{Vision} accuracy changes by at most $1.33$ percentage points, with no statistically significant change, consistent with preserved performance on this control.
In \texttt{Standard}, InternVL3-8B improves by $6.42$ and $7.17$ percentage points after GSM8K and MMLU self-distillation, respectively (both $p<10^{-4}$).

\vspace{-1mm}
\mypara{Ablations: filtering and adaptation strategy.}
We then ablate two design choices on Qwen3-VL-8B + GSM8K (Table~\ref{tab:selfdistillablation}).
First, we compare \emph{filtered} traces (only correct answers) against \emph{unfiltered} traces (all outputs regardless of correctness): the gap is only 1.29\%, suggesting that, as long as the model maintains a good level of textual reasoning capability, the reasoning \emph{structure} matters more than perfect correctness, showing the potential to conduct self-distillation with unlabeled data.
\begin{wraptable}{r}{0.52\textwidth}
\centering
\scriptsize
\setlength{\tabcolsep}{3pt}
\begin{tabular}{@{}l cc cc c@{}}
\toprule
& \multicolumn{2}{c}{\textbf{Baseline}} & \multicolumn{2}{c}{\textbf{Self-Distilled}} & \\
\cmidrule(lr){2-3} \cmidrule(lr){4-5}
\textbf{Strategy} & \makecell{\scriptsize\textbf{Text}\\\scriptsize(\TextCircle)} & \makecell{\scriptsize\textbf{Image}\\\scriptsize(\ImageCircle)} & \makecell{\scriptsize\textbf{Text}\\\scriptsize(\TextCircle)} & \makecell{\scriptsize\textbf{Image}\\\scriptsize(\ImageCircle)} & \makecell{\scriptsize$|\Delta|$} \\
\midrule
ViT+LM (unfiltered) & \multirow{4}{*}{93.56} & \multirow{4}{*}{30.71} & 93.71 & 91.28 & 2.43 \\
ViT+LM (filtered) &  &   & 93.03 & 92.57 & \textbf{0.46} \\
ViT-only &  &  & 93.25 & 85.29 & 7.96 \\
LM-only &  &  & \textbf{94.09} & \textbf{92.72} & 1.37 \\
\bottomrule
\end{tabular}
\vspace{1mm}
\caption{Ablation on Qwen3-VL-8B + GSM8K: filtered vs.\ unfiltered traces, and ViT+LM / ViT-only / LM-only adaptation. LM-only with filtered traces is the canonical recipe. $|\Delta|$: text--image gap.}
\label{tab:selfdistillablation}
\vspace{-12pt}
\end{wraptable}

Second, we compare three adaptation strategies: ViT+LM (Full LoRA), LM-only (freeze ViT), and ViT-only (freeze LM).
ViT-only training raises image-mode accuracy to 85.29\%, showing that better visual feature alignment alone yields substantial gains.
LM-only training achieves 92.72\%, matching ViT+LM (92.57\%), indicating that adapting the language model is the key factor, reducing the effort from tuning the full model to tuning only the LM component.

\vspace{-2mm}
\section{Conclusion}
\vspace{-2mm}
We move beyond reporting the text-vs-image accuracy gap to diagnosing why it arises and how it can be closed. 
Across seven benchmarks, seven models, and five input modes, three findings reshape how the modality gap should be understood. First, MLLMs already perform strongly on natural document images and struggle primarily on synthetic renderings whose visual properties diverge from pretraining distributions, with font, resolution, and compression alone shifting accuracy by tens of percentage points.
Second, an error taxonomy built from over 5,000 failures reveals that image mode does not impair perception or knowledge retrieval but triggers a reasoning collapse in which models produce shorter outputs and skip step-by-step reasoning.
The reluctance to reason, not the inability to read, drives the gap. 
Third, this diagnosis suggests a lightweight remedy: self-distillation on the model's own text-mode reasoning traces, paired with image inputs, closes the gap without architectural changes, external data, or catastrophic forgetting.
These findings translate into concrete guidance for two audiences: practitioners using existing MLLMs should add explicit chain-of-thought prompts to visual inputs and attend to rendering choices. 
Model builders, in turn, should treat long-form reasoning over visual inputs as a major training objective rather than relying on short VQA-style answers.
Together, our results suggest that the reported modality gaps stem less from a fundamental limit of pixel-based language understanding than from a fixable misalignment between how MLLMs are trained to behave on visual inputs and how they are capable of reasoning over them.

\begin{ack}
Part of this study is supported by the Lambda Research Grant Program by lambda.ai.
K.S. was supported in part by DARPA (SciFY). The U.S.
Government is authorized to reproduce and distribute reprints for Governmental purposes. The views and conclusions contained in this publication are those of the authors and should not be interpreted as representing official policies or endorsements of DARPA or the U.S. Government.
\end{ack}

\bibliographystyle{unsrt}
\bibliography{custom}

\appendix

\setcounter{page}{1}

\setcounter{figure}{0}
\setcounter{table}{0}
\renewcommand\thesection{\Alph{section}}
\renewcommand{\thetable}{S\arabic{table}}  
\renewcommand{\thefigure}{S\arabic{figure}}

\appendix

\section{Realistic Data Preparation}
\textbf{\textsc{Qasper}} \cite{Dasigi2021ADO} is a long-form QA dataset over scientific articles that often requires multi-sentence reasoning.
All instances that require access to tables and figures are removed to ensure a fair comparison between the \texttt{Pure Text} format and \texttt{Pure Image} format.
We download the original PDF files from arXiv and convert all pages into PNG images using the \textit{fitz} library, excluding the reference section, which is also not included in the \texttt{Pure Text} format of QASPER.
For each example in \textbf{SQuAD}~v2,  we capture a screenshot of the corresponding Wikipedia page to create a natural-image version of the context. 
The corresponding screenshot is manually aligned with the passages given in the original SQuAD~v2 data.
Due to the nature of webpages, there would be unavoidable truncations of images or tables in the \texttt{Pure Image} input that is not included by the \texttt{Pure Text} setting.
For evaluation, zero-shot inference on QA tasks does not align well with the original span-based metrics, because MLLMs are not trained to reproduce annotation artifacts such as exact spans. Instead, following recent information-seeking benchmarks \citep{wei2025browsecompsimplechallengingbenchmark, pham2025sealqaraisingbarreasoning}, we adopt an LLM-as-judge protocol, where use GPT-5 (\textit{gpt-5-2025-08-07}) to compare the model prediction with the gold answer and report accuracy over all examples. Results on span-based metrics can be found in \Cref{app:evaluation}.

\section{Detailed Input Settings}
\label{app:input-configs}

\begin{figure}[h] 
  \centering
  \begin{subfigure}{0.9\columnwidth}
    \centering
    \includegraphics[width=0.75\linewidth,keepaspectratio]{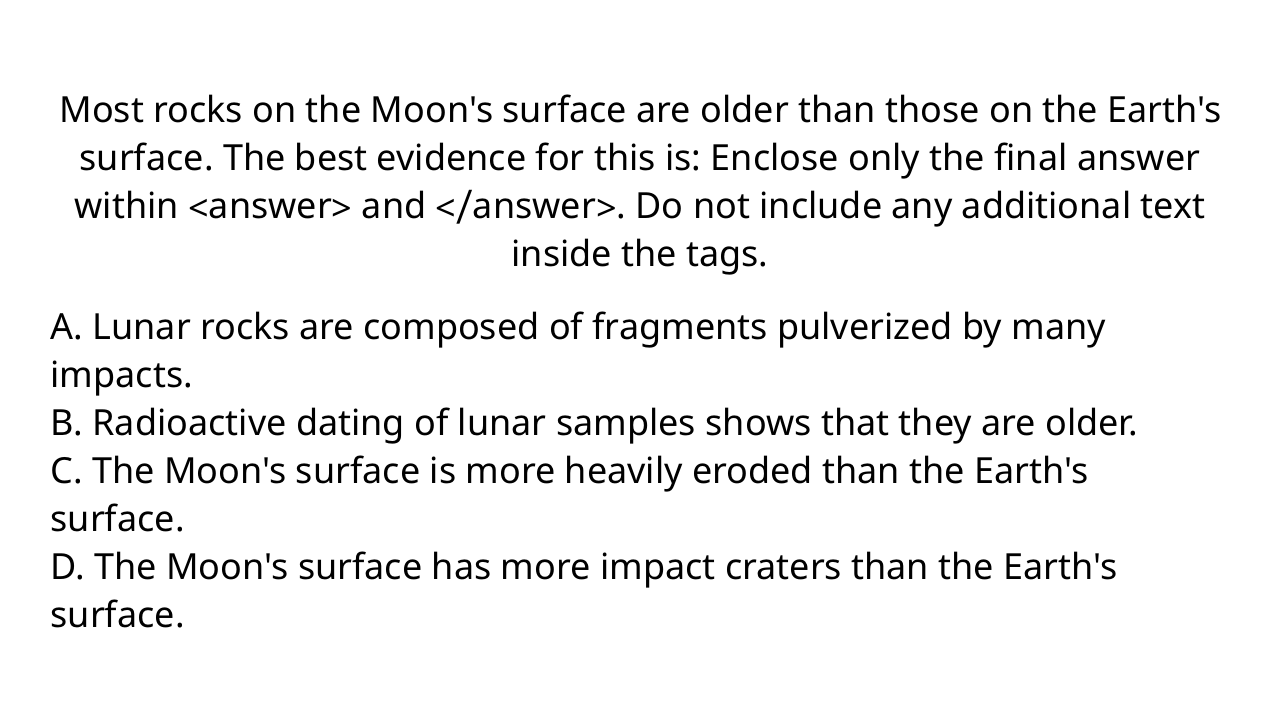} %
    \caption{\texttt{Image} format.}
    \label{fig:mmlu-example}
  \end{subfigure}

  \vspace{0.8em}

\begin{subfigure}{\columnwidth}
  \begin{fullwidthExampleBoxColumn}
    \begin{tabularx}{\linewidth}{@{}p{0.12\linewidth}X@{}}
      \textbf{Text}  & Answer the following multiple choice question. Respond with just the letter of the correct answer (A, B, C, or D).
      
    Most rocks on the Moon's surface are older than those on the Earth's surface. The best evidence for this is:

    A. Lunar rocks are composed of fragments pulverized by many impacts.
    
    B. Radioactive dating of lunar samples shows that they are older.
    
    C. The Moon's surface is more heavily eroded than the Earth's surface.
    
    D. The Moon's surface has more impact craters than the Earth's surface.
    
    Enclose only the final answer within <answer> and </answer>. Do not include any additional text inside the tags.
    \\
    \textbf{Instr. + Image} & Please follow the instruction in the image. Enclose only the final answer within <answer> and </answer>. Do not include any additional text inside the tags. [Image provided in (a) without the format instructions.]
     \\
    \end{tabularx}
  \end{fullwidthExampleBoxColumn}
  \caption{\texttt{Text} and \texttt{Instr. + Image} format.}
    \label{app:tab:mmlu-example-others}
   \end{subfigure}
    
  \caption{An example input of MMLU.}
  \label{tab:mmluinput}
\end{figure}

This section provides the full input formats used across all configurations, along with examples for each dataset.
Unless otherwise specified, models are prompted to produce outputs wrapped in \texttt{<answer>...</answer>}. 
For datasets that require structured or non–free-form outputs, we apply a permissive parsing strategy to extract the final predictions.
All evaluations are conducted in a zero-shot setting. 
Accuracy is used as the primary metric for all benchmarks, with the exception of HumanEval, which is evaluated using pass@k, where $k=1$.
\paragraph{MMLU}
MMLU evaluates a model’s knowledge and reasoning ability across 57 subjects spanning the humanities, social sciences, STEM fields, and professional domains. 
Each question is multiple-choice and often requires domain-specific expertise. 
In our setting, we adopt its original tasks and targets.
The example input for MMLU is shown in \Cref{tab:mmluinput}.

\newtcolorbox{fullwidthExampleBox}[1][]{%
  enhanced, 
  colback=gray!5,    colframe=gray!40,
  sharp corners,     boxrule=0.4pt,
  left=3pt, right=3pt, top=4pt, bottom=4pt,
  width=\textwidth,  %
  fontupper=\scriptsize\ttfamily,
  title style={font=\tiny\bfseries},
  #1}

\newtcolorbox{fullwidthExampleBoxColumn}[1][]{%
  enhanced,
  colback=gray!5,
  colframe=gray!40,
  sharp corners,
  boxrule=0.4pt,
  left=3pt, right=3pt, top=4pt, bottom=4pt,
  width=\linewidth,     %
  fontupper=\scriptsize\ttfamily,
  title style={font=\tiny\bfseries},
  #1
}

\begin{figure}[h] 
  \centering
  \begin{subfigure}{0.9\columnwidth}
    \centering
    \includegraphics[width=0.75\linewidth,keepaspectratio]{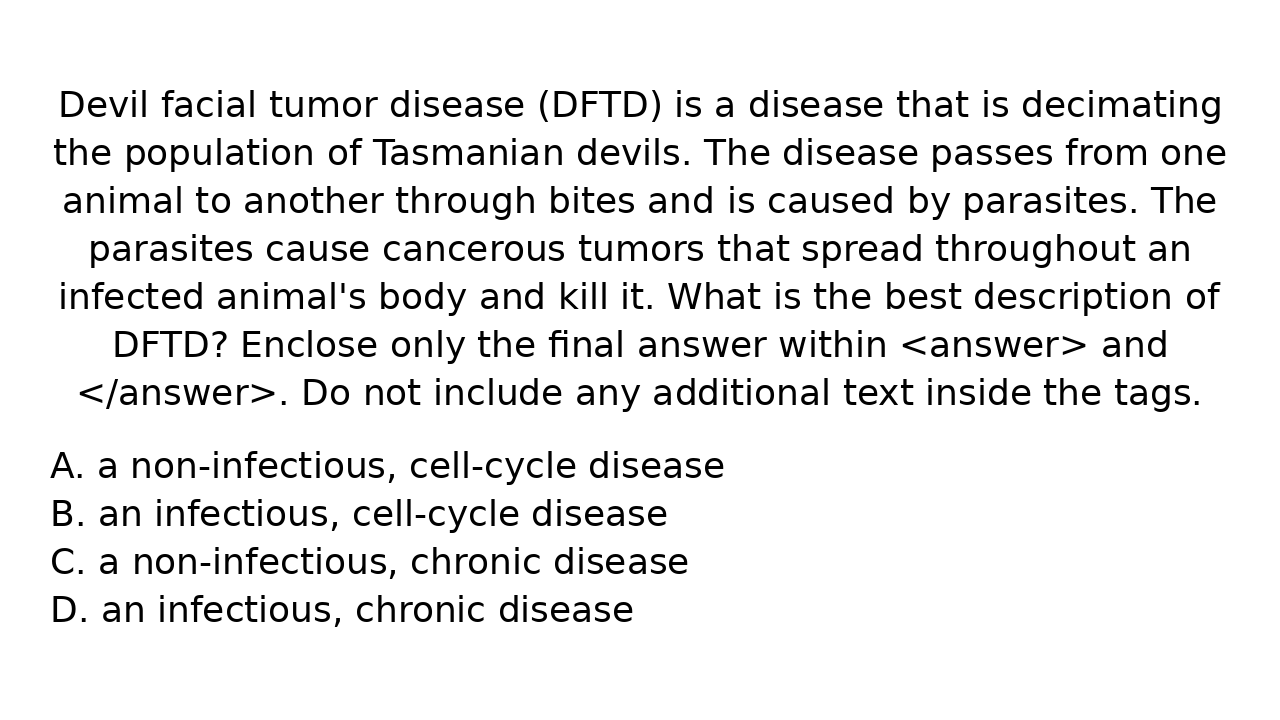} %
    \caption{\texttt{Image} format.}
    \label{fig:arc-example}
  \end{subfigure}

  \vspace{0.8em}

  \begin{subfigure}{\columnwidth}
  \begin{fullwidthExampleBoxColumn}
    \begin{tabularx}{\linewidth}{@{}p{0.12\linewidth}X@{}}
      \textbf{Text}  & Answer the following multiple choice question. Respond with just the letter of the correct answer (A, B, C, or D). 
      
      Devil facial tumor disease (DFTD) is a disease that is decimating the population of Tasmanian devils. The disease passes from one animal to another through bites and is caused by parasites. The parasites cause cancerous tumors that spread throughout an infected animal's body and kill it. What is the best description of DFTD?

      A. a non-infectious, cell-cycle disease", "an infectious, cell-cycle disease

      B. an infectious, cell-cycle disease

      C. a non-infectious, chronic disease

      D. an infectious, chronic disease

      Enclose only the final answer within <answer> and </answer>. Do not include any additional text inside the tags.
    \\
    \textbf{Instr. + Image} & Please follow the instruction in the image. Enclose only the final answer within <answer> and </answer>. Do not include any additional text inside the tags. [Image provided in (a) without the format instructions.]
     \\
    \end{tabularx}
  \end{fullwidthExampleBoxColumn}
  \caption{\texttt{Text} and \texttt{Instr. + Image} format.}
    \label{app:tab:arc-example-others}
   \end{subfigure}
   
  \caption{An example input of ARC.}
  \label{tab:arcinput}
\end{figure}

\paragraph{ARC}
ARC-Challenge (ARC) assesses a model’s ability to perform complex, non-trivial reasoning on grade-school science questions.
Unlike its easier counterpart (ARC-Easy), ARC is intentionally adversarial: questions cannot be solved by surface heuristics and typically require multi-step inference, commonsense knowledge, or integration of scientific facts.
We adopt its original tasks and targets, and the example input for ARC-C is shown in \Cref{tab:arcinput}.

\begin{figure}[h]
  \centering
  \begin{subfigure}{0.9\columnwidth}
    \centering
    \includegraphics[width=0.75\linewidth,keepaspectratio]{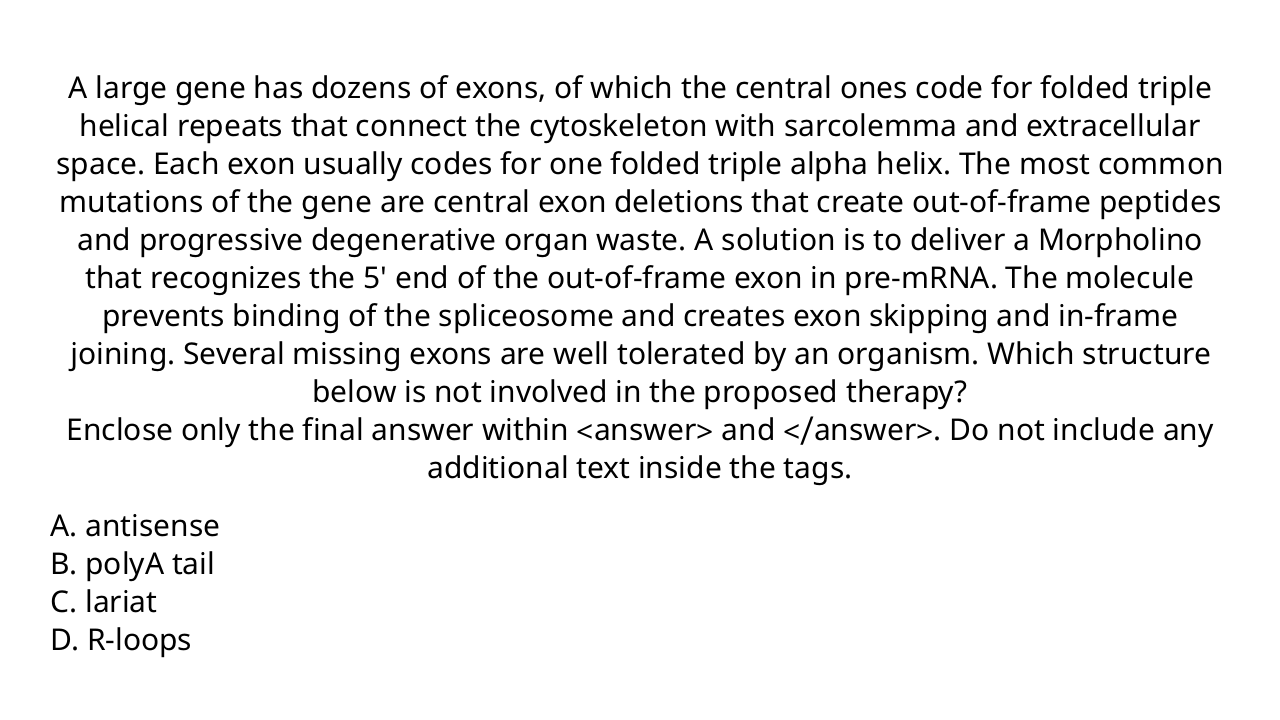}
    \caption{\texttt{Image} format.}
    \label{fig:gpqa-example}
  \end{subfigure}

  \vspace{0.8em}

  \begin{subfigure}{\columnwidth}
  \begin{fullwidthExampleBoxColumn}
    \begin{tabularx}{\linewidth}{@{}p{0.12\linewidth}X@{}}
      \textbf{Text}  & 
      A large gene has dozens of exons, of which the central ones code for folded triple helical repeats that connect the cytoskeleton with sarcolemma and extracellular space. Each exon usually codes for one folded triple alpha helix. The most common mutations of the gene are central exon deletions that create out-of-frame peptides and progressive degenerative organ waste. A solution is to deliver a Morpholino that recognizes the 5' end of the out-of-frame exon in pre-mRNA. The molecule prevents binding of the spliceosome and creates exon skipping and in-frame joining. Several missing exons are well tolerated by an organism. Which structure below is not involved in the proposed therapy?

    Enclose only the final answer within <answer> and </answer>. Do not include any additional text inside the tags.

    A. antisense
    B. polyA tail
    C. lariat
    D. R-loops
    \\
    \textbf{Instr. + Image} & Please follow the instruction in the image. Enclose only the final answer within <answer> and </answer>. Do not include any additional text inside the tags. [Image provided in (a) without the format instructions.]
     \\
    \end{tabularx}
  \end{fullwidthExampleBoxColumn}
  \caption{\texttt{Text} and \texttt{Instr. + Image} format.}
    \label{app:tab:gpqa-example-others}
   \end{subfigure}

  \caption{An example input of GPQA.}
  \label{tab:GPQAinput}
\end{figure}

\paragraph{GPQA}
GPQA \cite{rein2023gpqa} is a graduate-level question answering benchmark consisting of 448 challenging multiple-choice questions in physics, chemistry, and biology.
The questions are crafted by domain experts and are designed to be difficult even for specialists outside their own subfield, requiring deep domain knowledge and multi-step scientific reasoning.
We adopt its original tasks and targets, and the example input for GPQA is shown in \Cref{tab:GPQAinput}.

\begin{figure}[h] 
  \centering
  \begin{subfigure}{\columnwidth}
    \centering
    \includegraphics[width=0.75\linewidth,keepaspectratio]{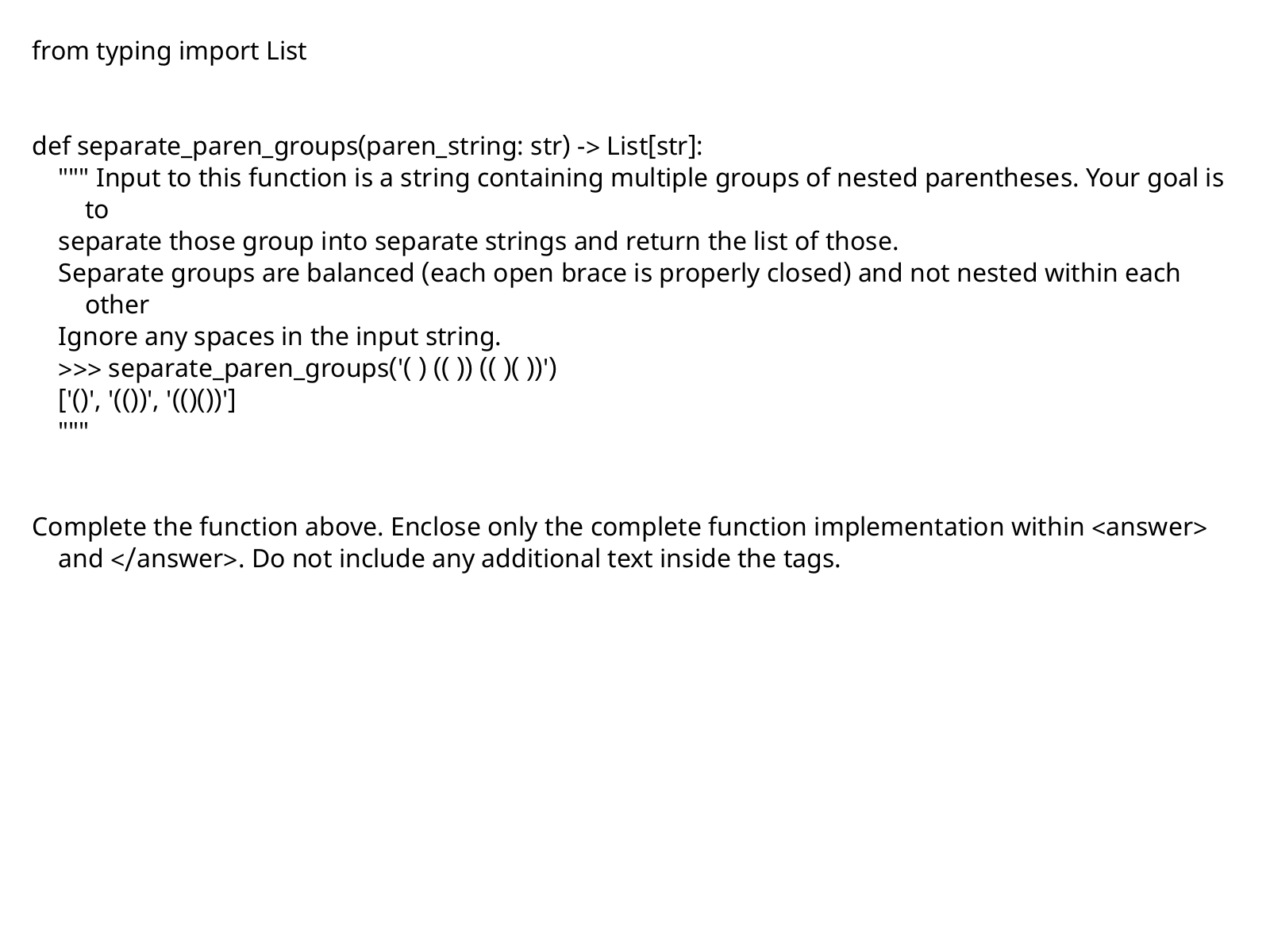} %
    \caption{\texttt{Image} format.}
    \label{fig:HumanEval-example}
  \end{subfigure}

  \vspace{0.8em}

    \begin{subfigure}{\columnwidth}
  \begin{fullwidthExampleBoxColumn}
    \begin{tabularx}{\textwidth}{@{}p{0.12\textwidth}X@{}}  %

      \textbf{Text} &
      \begin{minipage}[t]{\linewidth}
        \ttfamily
from typing import List

def separate\_paren\_groups(paren\_string: str) -> List[str]:
    """ Input to this function is a string containing multiple groups of nested
    parentheses. Your goal is to
    separate those group into separate strings and return the list of those.
    Separate groups are balanced (each open brace is properly closed) and not
    nested within each other
    Ignore any spaces in the input string.
    >>> separate\_paren\_groups('( ) (( )) (( )( ))')
    ['()', '(())', '(()())']
    """

        \medskip

        Enclose only the final answer within \texttt{<answer>} and \texttt{</answer>}.
        Do not include any additional text inside the tags.
      \end{minipage}
      \\[0.75em]

      \textbf{Instr. + Image} &
      Please follow the instruction in the image.
      Enclose only the final answer within \texttt{<answer>} and \texttt{</answer>}.
      Do not include any additional text inside the tags. [Image provided in (a) without the format instructions.]
      \\

    \end{tabularx}
  \end{fullwidthExampleBoxColumn}
  \caption{\texttt{Text} and \texttt{Instr. + Image} format.}
  \label{app:tab:humanEval-example-others}
\end{subfigure}

  \caption{An example input of HumanEval.}
  \label{tab:HumanEvalinput}
\end{figure}

\paragraph{HumanEval}
HumanEval \cite{chen2021evaluating} measures a model’s ability to synthesize correct and executable code from natural language descriptions.
Each task provides a short docstring specifying a function’s intended behavior, and models must generate a complete Python implementation that passes hidden unit tests.
The example input of HumanEval is shown in \Cref{tab:HumanEvalinput}.

\paragraph{GSM8K}

\begin{figure}[h] 
  \centering
  \begin{subfigure}{0.9\columnwidth}
    \centering
    \includegraphics[width=0.55\linewidth,keepaspectratio]{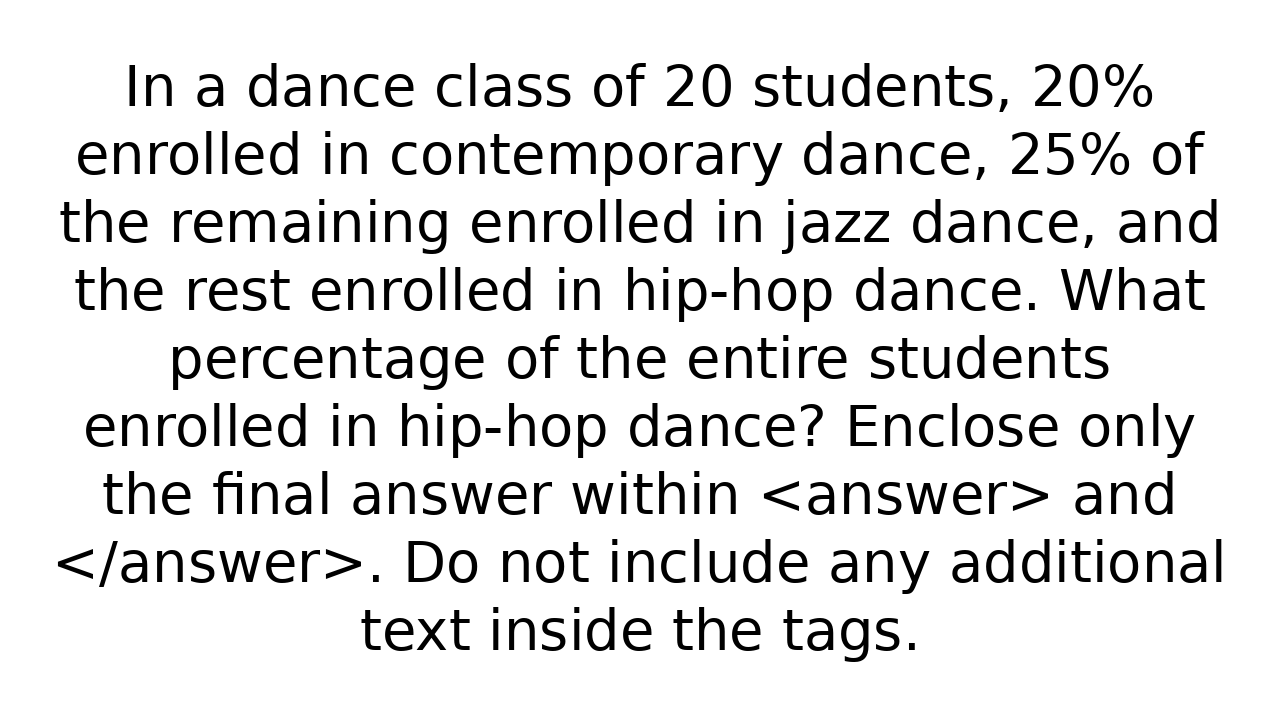} %
    \caption{\texttt{Image} format.}
    \label{fig:GSM8K-example}
  \end{subfigure}

  \vspace{0.8em}

  \begin{subfigure}{\columnwidth}
  \begin{fullwidthExampleBoxColumn}
    \begin{tabularx}{\columnwidth}{@{}p{0.12\textwidth}X@{}}  %
      \textbf{Text}  & 
      In a dance class of 20 students, 20\% enrolled in contemporary dance, 25\% of the remaining enrolled in jazz dance, and the rest enrolled in hip-hop dance. What percentage of the entire students enrolled in hip-hop dance?

      Enclose only the final answer within <answer> and </answer>. Do not include any additional text inside the tags.
    \\
    \textbf{Instr. + Image} & Please follow the instruction in the image. Enclose only the final answer within <answer> and </answer>. Do not include any additional text inside the tags. [Image provided in (a) without the format instructions.]
     \\
    \end{tabularx}
  \end{fullwidthExampleBoxColumn}
  \caption{\texttt{Text} and \texttt{Instr. + Image} format.}
    \label{app:tab:gsm8k-example-others}
   \end{subfigure}
  \caption{An example input of GSM8K.}
  \label{tab:GSM8Kinput}
\end{figure}

GSM8K evaluates a model's ability to solve grade-school mathematical word problems that require multi-step reasoning.
Each question is free-form and demands 2 to 8 steps to arrive at the final answer.
The solution of GSM8K is provided in natural language format, and a final answer is given after the reasoning steps.
In our setting, we adopt its original data and consider primarily the correctness of the answer (extracting the answer from the reasoning chain).
The example input for GSM8K is shown in \Cref{tab:GSM8Kinput}.
To avoid confounding factors introduced by in-context examples, all the experiments are run on the zero-shot condition.

\paragraph{SQuAD v2.}
SQuAD v2 is an extractive question answering dataset that tests a model’s ability to extract the answers from a given passage and to recognize unanswerable queries.
Each instance presents a question and a supporting passage from Wikipedia, and models must either extract a precise span or correctly abstain when the passage lacks sufficient information.

\begin{figure}[ht] 
  \centering
  \begin{subfigure}{\columnwidth}
    \centering
    \includegraphics[width=0.7\linewidth,keepaspectratio]{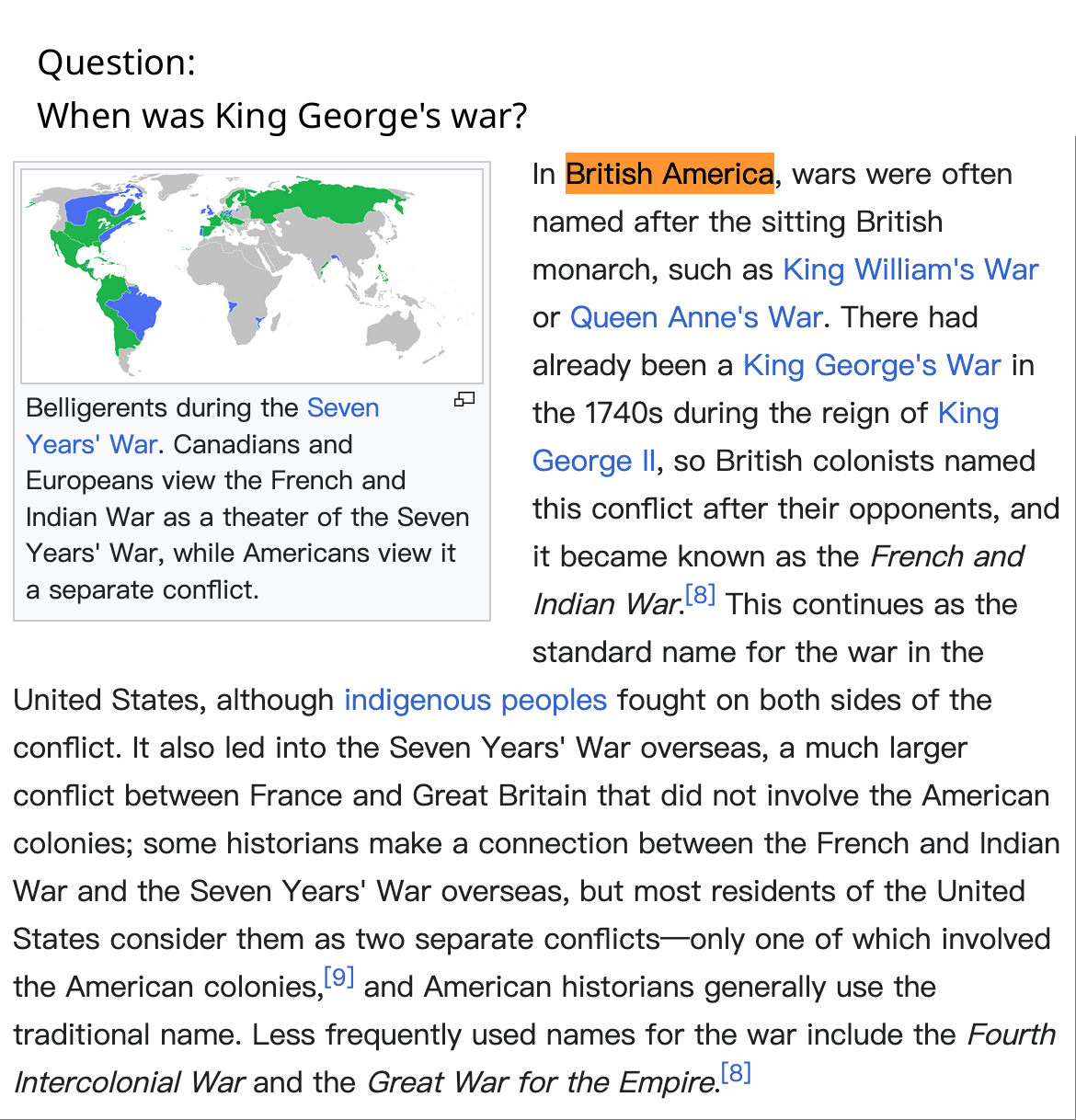} %
    \caption{Example \texttt{Image} input from a Wikipedia screenshot, preserving the webpage's typography, layout, and visual elements. The captured page may differ from the original SQuAD passage.}
    \label{fig:SQUAD-example}
  \end{subfigure}

  \vspace{0.8em}

  \begin{subfigure}{\columnwidth}
  \begin{fullwidthExampleBoxColumn}
    \begin{tabularx}{\textwidth}{@{}p{0.12\textwidth}X@{}}  %
      \textbf{Text}  & 
      When was King George's war?
The conflict is known by multiple names. In British America, wars were often named after the sitting British monarch, such as King William's War or Queen Anne's War. As there had already been a King George's War in the 1740s, British colonists named the second war in King George's reign after their opponents, and it became known as the French and Indian War. This traditional name continues as the standard in the United States, but it obscures the fact that Indians fought on both sides of the conflict, and that this was part of the Seven Years' War, a much larger conflict between France and Great Britain. American historians generally use the traditional name or sometimes the Seven Years' War. Other, less frequently used names for the war include the Fourth Intercolonial War and the Great War for the Empire. 
      Enclose only the final answer within <answer> and </answer>. Do not include any additional text inside the tags.
    \\
    \textbf{Instr. + Image} & Please follow the instructions in the image. Enclose only the final answer within <answer> and </answer>. Do not include any additional text inside the tags. [Image provided in (a) without the format instructions.]
     \\
    \end{tabularx}
  \end{fullwidthExampleBoxColumn}
  \caption{\texttt{Text} and \texttt{Instr. + Image} format.}
    \label{app:tab:squad-example-others}
   \end{subfigure}

  \caption{An example input of SQUAD v2.}
  \label{tab:SQUADinput}
\end{figure}

\paragraph{QASPER.}

\begin{figure*}[t] 
  \centering
  \begin{subfigure}{0.9\textwidth}
    \centering
    \includegraphics[width=0.5\linewidth,keepaspectratio]{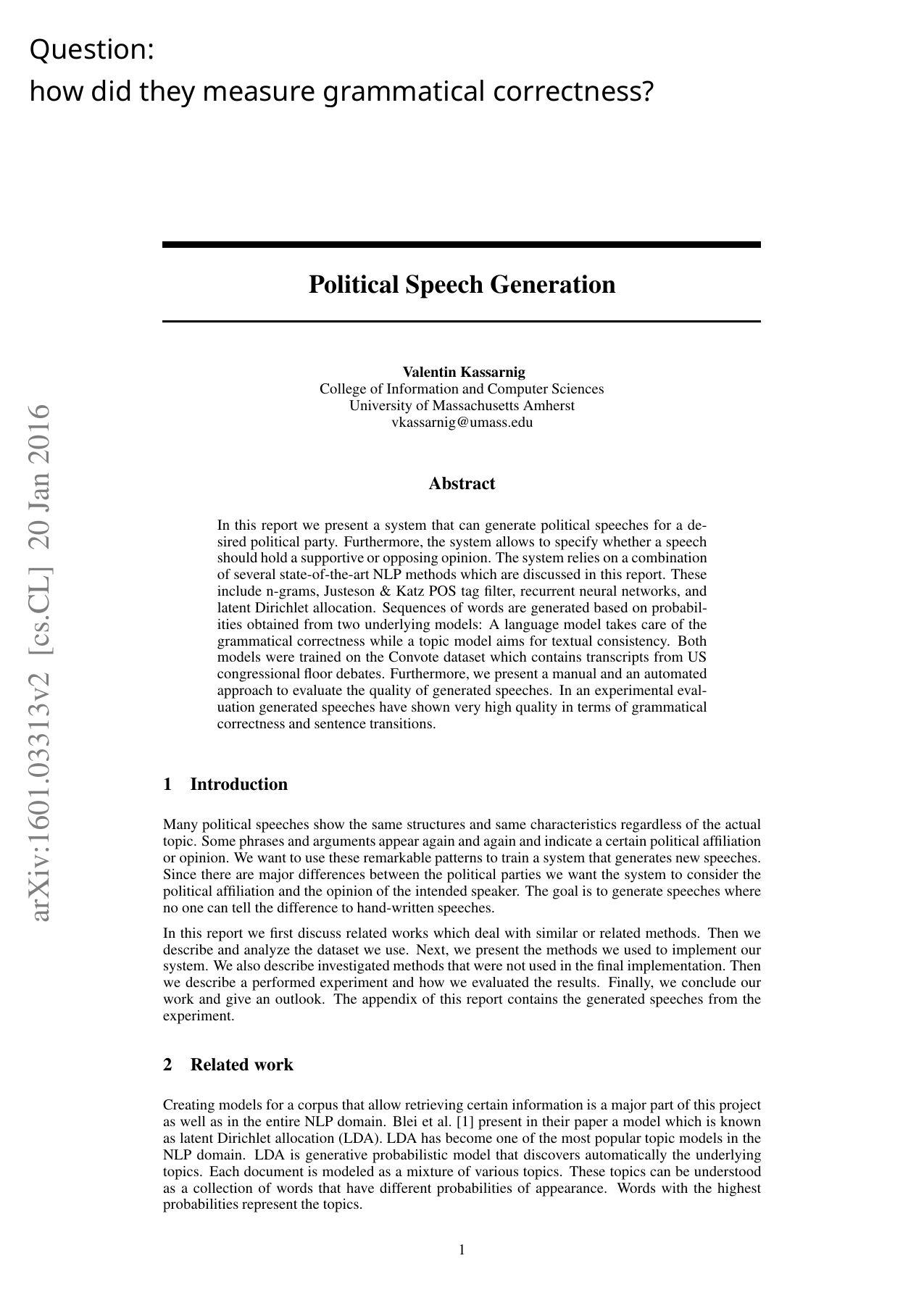} %
    \caption{The first image of \texttt{Image} format. The remaining pages are omitted for length considerations. All pages of the paper are fed into the model.}
    \label{fig:QASPER-example}
  \end{subfigure}

  \vspace{0.8em}

  \begin{subfigure}{0.9\textwidth}
  \begin{fullwidthExampleBox}
    \begin{tabularx}{\textwidth}{@{}p{0.06\textwidth}X@{}}  %
      \textbf{Text}  & 
      Question: how did they measure grammatical correctness?
      
      Political Speech Generation
      
      In this report we present a system that can generate political speeches for a desired political party. Furthermore, the system allows to specify whether a speech should hold a supportive or opposing opinion. The system relies on a combination of several state-of-the-art NLP methods which are discussed in this report. These include n-grams, ...
      ...[Full Paper Text]...

      Enclose only the final answer within <answer> and </answer>. Do not include any additional text inside the tags.
    \\
    \textbf{Instr. + Image} & Please follow the instructions in the image. Enclose only the final answer within <answer> and </answer>. Do not include any additional text inside the tags. [Image provided in (a) without the format instructions.]
     \\
    \end{tabularx}
  \end{fullwidthExampleBox}
  \caption{\texttt{Text} and \texttt{Instr. + Image} format.}
    \label{app:tab:qasper-example-others}
   \end{subfigure}

  \caption{An example input of QASPER.}
  \label{tab:QASPERinput}
\end{figure*}

QASPER is a long-context reading comprehension task that evaluates a model’s ability to answer questions about academic research papers.
Each question targets specific information regarding methods, findings, assumptions, or limitations of the paper, requiring models to reason over evidence within full-length scientific documents.
We feed the entire paper on a given question as the input.
The example input for QASPER is shown in \Cref{tab:QASPERinput}.

\subsection{OCR Prompt}
The prompt used in OCR-P2 is shown in \Cref{fig:ocr-prompt}.
It is also used in OCR-P1 as the guidelines for performing extractions.

\begin{figure*}[h]
  \centering
  \begin{tcolorbox}[
      enhanced,
      title={\normalsize OCR Extraction Prompt.},
      colback=gray!3,
      colframe=gray!60,
      left=1mm,right=1mm,
      top=0.6em,bottom=0.6em,
      boxrule=0.4pt,
      listing only,
      listing options={
         basicstyle=\ttfamily\small,
         breaklines=true,
         breakatwhitespace=true,
         columns=fullflexible,
      }
    ]
You are an Optical Character Recognition (OCR) model. Your task is to extract all readable text from the provided image.

Instructions:

1. Identify and transcribe all visible text exactly as it appears, including numbers, punctuation, and symbols.

2. Preserve line breaks and spacing whenever possible.

3. For structured content (tables, forms, equations, lists), format using simple text notation:

   \text{        }- Tables: use tabs or | separators between columns
   
   \text{        }- Lists: use bullet points (-) or numbers
   
   \text{        }- Equations: reproduce mathematical symbols faithfully
   
4. If a word or section is unclear, write [illegible] in its place.

5. Do not interpret, summarize, or translate the text—transcribe it exactly as shown.
\end{tcolorbox}

  \caption{Prompt used for OCR-style transcription from images.}
  \label{fig:ocr-prompt}
\end{figure*}

\section{Dataset Statistics}
\Cref{tab:dataset_stat} summarizes the composition of our evaluation suite.
The five synthetic-image benchmarks (MMLU, ARC, GPQA, GSM8K, and HumanEval) comprise 17,145 instances in total, with MMLU accounting for the largest share (81.9\%).
QASPER and SQuAD v2 use natural images and are listed separately.
\begin{table}[ht]
\centering
\resizebox{0.8\columnwidth}{!}{
\begin{tabular}{lcccc}
\toprule
\textbf{Dataset} & \textbf{Domain} & \textbf{Metric} & \textbf{\#Samples} & \textbf{Proportion (\%)} \\
\midrule
MMLU          & Knowledge (57 subjects) & Acc.       & 14,042 & 81.9 \\
ARC           & Science reasoning       & Acc.       & 1,172  & 6.8  \\
GPQA          & Graduate-level QA       & Acc.       & 448    & 2.6  \\
GSM8K         & Math reasoning          & Acc.       & 1,319  & 7.7  \\
HumanEval     & Code generation         & Pass$@$1   & 164    & 1.0  \\
\hdashline
QASPER        & Document QA             & F1 / Acc.  & 1,445  & ---  \\
SQuAD v2      & Reading comprehension   & F1 / Acc.  & 100    & ---  \\
\midrule
\multicolumn{3}{l}{\textbf{Total (synthetic-image benchmarks)}} & \textbf{17,145} & \textbf{100.0} \\
\bottomrule
\end{tabular}
}
\vspace{8pt}
\caption{Dataset statistics and composition.}
\label{tab:dataset_stat}
\end{table}

\section{Experiment Setting Details}

We use a unified decoding setup across models and datasets.
Unless otherwise noted, the temperature is set to 0.1 with a batch size of 8 and a maximum generation length of 1024 tokens.
For code generation tasks, the temperature is set to 0.2 to ensure diversity on the generation results.

For OCR-1P, the maximum generation length is set to 4096, as the model is asked to perform OCR and solve the task in one inference.
The generation length is increased to allow full inclusion of OCR results and task reasoning in one output.

\section{HumanEval Results}
\label{app:humaneval}

We report HumanEval results separately due to the high variance inherent in this benchmark.
With only 164 instances, pass@1 estimates are statistically unstable; Chen et al.~\citep{chen2021evaluating} recommend generating many samples per problem and using an unbiased estimator for reliable evaluation.
Additionally, code generation is particularly sensitive to formatting failures (Section~\ref{sec:error_analysis}).

\begin{table}[ht]
\centering
\scriptsize
\begin{tabular}{@{}l|ccccc@{}}
\toprule
\textbf{Model} & \makecell{\scriptsize\textbf{Pure}\\\scriptsize\textbf{Text}\\\scriptsize(\TextCircle)} & \makecell{\scriptsize\textbf{Instr.+}\\\scriptsize\textbf{Image}\\\scriptsize(\TextCircle\ImageCircle)} & \makecell{\scriptsize\textbf{Pure}\\\scriptsize\textbf{Image}\\\scriptsize(\ImageCircle)} & \makecell{\scriptsize\textbf{OCR-1P}\\\scriptsize(\ImageCircle→\TextCircle)} & \makecell{\scriptsize\textbf{OCR-2P}\\\scriptsize(\ImageCircle→\TextCircle)} \\
\midrule
\tiny{GPT-5.2} & 100.00 & \cellcolor{blue!29}{71.34}$^{\dagger}$ & 100.00 & \cellcolor{blue!7}{98.17} & 100.00 \\
\tiny{InternVL3-8B} & 56.71 & \cellcolor{red!18}{71.95} & \cellcolor{blue!15}{45.12} & \cellcolor{red!17}{71.34} & \cellcolor{red!23}{78.05}$^{\dagger}$ \\
\tiny{InternVL3.5-8B} & 90.24 & 90.24 & \cellcolor{blue!6}{89.02} & \cellcolor{blue!10}{84.76} & \cellcolor{blue!8}{86.59} \\
\tiny{Pixtral-12B} & 39.02 & \cellcolor{red!39}{79.27}$^{\dagger}$ & \cellcolor{red!12}{47.56} & \cellcolor{red!16}{52.44} & \cellcolor{blue!38}{0.00}$^{\dagger}$ \\
\tiny{Qwen2.5-7B-VL} & 59.15 & \cellcolor{red!25}{83.54}$^{\dagger}$ & \cellcolor{red!10}{65.24} & \cellcolor{blue!11}{52.44} & \cellcolor{red!9}{64.02} \\
\tiny{Qwen2.5-32B-VL} & 69.51 & \cellcolor{blue!44}{22.56} & \cellcolor{red!19}{85.98} & \cellcolor{red!18}{84.76} & \cellcolor{red!16}{82.32} \\
\tiny{Qwen3-VL-8B} & 76.22 & \cellcolor{red!11}{82.93} & \cellcolor{red!11}{82.93} & \cellcolor{red!13}{85.37} & \cellcolor{red!14}{87.20} \\
\bottomrule
\end{tabular}
\vspace{2mm}
\caption{HumanEval (pass@1) results across input modalities. Cell shading indicates performance difference relative to \texttt{Pure Text}: \colorbox{red!30}{red} = outperforms text, \colorbox{blue!30}{blue} = underperforms text. Results exhibit high variance due to the small dataset size (164 instances) and the inherent instability of pass@1 estimation from limited samples~\citep{chen2021evaluating}. $^{\dagger}$: significant difference from Pure Text (Holm-adjusted $p<0.05$; Appendix~\ref{app:significance}).}
\label{tab:humaneval_results}
\end{table}

\section{Rendering Significance Tests}
\label{app:rendering_significance}

For the rendering comparisons in Figure~\ref{fig:typefaceablation}, we pair each alternative rendering with the default on the same benchmark items.
We use two-sided exact McNemar tests for binary correctness outcomes and paired bootstrap tests for HumanEval, where the ten generated samples per problem in each alternative rendering are averaged before pairing with the single-sample default.
We use 10{,}000 paired bootstrap resamples for the confidence intervals and bootstrap tests, and apply Holm correction jointly across the 60 condition--default comparisons.

\section{Resolution Sensitivity}
\label{app:resolution_curves}

\begin{figure}[h]
    \centering
    \includegraphics[width=0.7\linewidth]{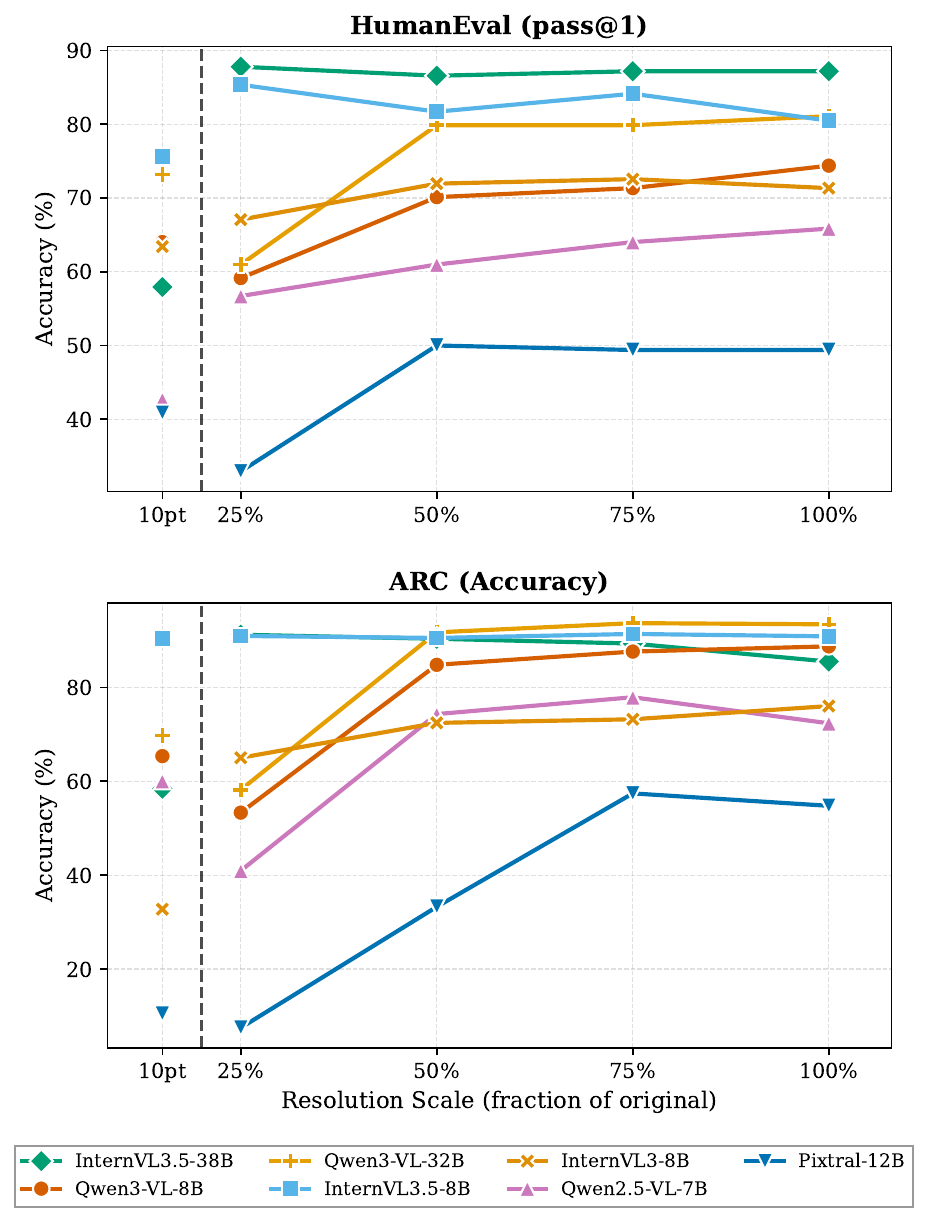}
    \caption{Performance vs.\ image resolution scale on HumanEval (top) and ARC (bottom). The black vertical dashed line indicates the point where \texttt{Pure Image} consumes the same FLOPs as \texttt{Pure Text}.}
    \label{fig:resolution_curves_full}
\end{figure}

Figure~\ref{fig:resolution_curves_full} presents the full resolution sensitivity analysis on both HumanEval and ARC.
Most models maintain stable performance until resolution reaches a lower bound, while InternVL3.5-8B maintains stable performance across all resolutions due to its Visual Resolution Router~\cite{wang2025internvl3}.

\section{Full Results Table}
\label{app:natural_rendered}
We report the complete performance of all models under all input configurations in \Cref{app:tab:fulltable}. 
The table covers \textit{PureText}, \textit{PureImage}, \textit{Inst.+Image} (Real vs.\ Rendered), and both OCR-based variants (OCR-1P and OCR-2P), evaluated across seven benchmarks: MMLU, ARC, GPQA, GSM8K, QASPER, SQuAD v2, and HumanEval. 
All results are reported as Accuracy or F1, except for \textbf{HumanEval}, which follows the standard \textbf{pass@1} metric. 
Accuracy values of QASPER and SQuAD v2 are computed using the model-based evaluator described in Section~\ref{app:evaluation}, ensuring consistent comparison across heterogeneous answer formats.

\clearpage
\begingroup
\footnotesize
\setlength{\tabcolsep}{3pt}
\begin{longtable}{@{}lrrrrrrr@{}}
\caption{Performance across input modalities on all seven datasets. Values are percentages; definitions and provenance are discussed in Appendix~\ref{app:natural_rendered}. For the five diagnostic benchmarks, $^{\dagger}$ marks a significant difference from Pure Text (Holm-adjusted $p<0.05$; Appendix~\ref{app:significance}).}\label{app:tab:fulltable}\label{tab:full_table_results}\\
\toprule
Model & Text & \makecell{Natural\\Instr.+Img.} & \makecell{Natural\\Pure Img.} & OCR-1P & OCR-2P & \makecell{Rendered\\Pure Img.} & \makecell{Rendered\\Instr.+Img.} \\
\midrule
\endfirsthead
\multicolumn{8}{l}{\tablename\ \thetable\ (continued)}\\
\toprule
Model & Text & \makecell{Natural\\Instr.+Img.} & \makecell{Natural\\Pure Img.} & OCR-1P & OCR-2P & \makecell{Rendered\\Pure Img.} & \makecell{Rendered\\Instr.+Img.} \\
\midrule
\endhead
\midrule
\multicolumn{8}{r}{Continued on next page}\\
\endfoot
\bottomrule
\endlastfoot
\multicolumn{8}{@{}l}{\textbf{MMLU (accuracy)}}\\*
GPT-5.2 & 92.33 & -- & -- & 86.54$^{\dagger}$ & 86.72$^{\dagger}$ & 90.93$^{\dagger}$ & 91.06$^{\dagger}$ \\
InternVL3-8B & 59.83 & -- & -- & 60.40 & 71.86$^{\dagger}$ & 33.93$^{\dagger}$ & 69.09$^{\dagger}$ \\
InternVL3.5-8B & 82.97 & -- & -- & 77.65$^{\dagger}$ & 80.75$^{\dagger}$ & 80.74$^{\dagger}$ & 82.25 \\
Pixtral-12B & 37.89 & -- & -- & 49.54$^{\dagger}$ & 44.25$^{\dagger}$ & 42.76$^{\dagger}$ & 41.60$^{\dagger}$ \\
Qwen2.5-7B-VL & 67.80 & -- & -- & 65.46$^{\dagger}$ & 67.08 & 67.36 & 67.97 \\
Qwen2.5-32B-VL & 78.34 & -- & -- & 76.55$^{\dagger}$ & 80.38$^{\dagger}$ & 73.79$^{\dagger}$ & 75.47$^{\dagger}$ \\
Qwen3-VL-8B & 78.44 & -- & -- & 72.98$^{\dagger}$ & 73.61$^{\dagger}$ & 71.01$^{\dagger}$ & 72.29$^{\dagger}$ \\
\midrule
\multicolumn{8}{@{}l}{\textbf{ARC (accuracy)}}\\*
GPT-5.2 & 96.25 & -- & -- & 97.70 & 97.18 & 95.48 & 92.24$^{\dagger}$ \\
InternVL3-8B & 92.24 & -- & -- & 86.26$^{\dagger}$ & 90.53 & 90.27 & 91.30 \\
InternVL3.5-8B & 92.66 & -- & -- & 93.26 & 93.52 & 92.66 & 92.41 \\
Pixtral-12B & 84.64 & -- & -- & 72.10$^{\dagger}$ & 39.33$^{\dagger}$ & 61.77$^{\dagger}$ & 71.25$^{\dagger}$ \\
Qwen2.5-7B-VL & 87.20 & -- & -- & 87.71 & 86.35 & 89.42 & 85.92 \\
Qwen2.5-32B-VL & 93.09 & -- & -- & 94.62 & 94.37 & 90.02$^{\dagger}$ & 91.30 \\
Qwen3-VL-8B & 91.72 & -- & -- & 92.92 & 92.75 & 91.04 & 91.38 \\
\midrule
\multicolumn{8}{@{}l}{\textbf{GPQA (accuracy)}}\\*
GPT-5.2 & 81.25 & -- & -- & 77.90 & 80.58 & 80.13 & 81.25 \\
InternVL3-8B & 33.71 & -- & -- & 30.58 & 33.26 & 32.81 & 32.14 \\
InternVL3.5-8B & 42.63 & -- & -- & 40.18 & 46.21 & 34.60$^{\dagger}$ & 40.40 \\
Pixtral-12B & 31.25 & -- & -- & 29.69 & 30.36 & 27.23 & 30.36 \\
Qwen2.5-7B-VL & 34.15 & -- & -- & 29.02 & 32.59 & 30.58 & 34.15 \\
Qwen2.5-32B-VL & 42.41 & -- & -- & 43.08 & 43.30 & 43.97 & 41.52 \\
Qwen3-VL-8B & 43.75 & -- & -- & 37.28 & 36.38 & 35.04$^{\dagger}$ & 39.29 \\
\midrule
\multicolumn{8}{@{}l}{\textbf{GSM8K (accuracy)}}\\*
GPT-5.2 & 95.83 & -- & -- & 77.86$^{\dagger}$ & 68.16$^{\dagger}$ & 96.51 & 85.14$^{\dagger}$ \\
InternVL3-8B & 78.01 & -- & -- & 73.92 & 87.87$^{\dagger}$ & 42.53$^{\dagger}$ & 49.51$^{\dagger}$ \\
InternVL3.5-8B & 95.30 & -- & -- & 95.60 & 94.47 & 95.30 & 95.22 \\
Pixtral-12B & 3.79 & -- & -- & 75.89$^{\dagger}$ & 55.12$^{\dagger}$ & 8.11$^{\dagger}$ & 4.93 \\
Qwen2.5-7B-VL & 19.48 & -- & -- & 20.09 & 20.17 & 24.79$^{\dagger}$ & 20.24 \\
Qwen2.5-32B-VL & 95.07 & -- & -- & 66.41$^{\dagger}$ & 96.13 & 94.92 & 4.47$^{\dagger}$ \\
Qwen3-VL-8B & 93.56 & -- & -- & 56.94$^{\dagger}$ & 94.77 & 30.71$^{\dagger}$ & 39.58$^{\dagger}$ \\
\midrule
\multicolumn{8}{@{}l}{\textbf{HumanEval (pass@1)}}\\*
GPT-5.2 & 100.00 & -- & -- & 98.17 & 100.00 & 100.00 & 71.34$^{\dagger}$ \\
InternVL3-8B & 56.71 & -- & -- & 71.34 & 78.05$^{\dagger}$ & 45.12 & 71.95 \\
InternVL3.5-8B & 90.24 & -- & -- & 84.76 & 86.59 & 89.02 & 90.24 \\
Pixtral-12B & 39.02 & -- & -- & 52.44 & 0.00$^{\dagger}$ & 47.56 & 79.27$^{\dagger}$ \\
Qwen2.5-7B-VL & 59.15 & -- & -- & 52.44 & 64.02 & 65.24 & 83.54$^{\dagger}$ \\
Qwen2.5-32B-VL & 69.51 & -- & -- & 84.76 & 82.32 & 85.98 & 22.56 \\
Qwen3-VL-8B & 76.22 & -- & -- & 85.37 & 87.20 & 82.93 & 82.93 \\
\midrule
\multicolumn{8}{@{}l}{\textbf{QASPER (full archive), F1}}\\*
GPT-5.2 & 29.18 & 29.36 & 28.28 & 28.89 & -- & 27.25 & 29.76 \\
InternVL3-8B & 21.60 & 16.32 & 12.51 & 15.51 & -- & 10.84 & 16.27 \\
Pixtral-12B & -- & -- & -- & -- & -- & -- & -- \\
Qwen2.5-32B-VL & 11.41 & 11.46 & 12.88 & 12.08 & -- & 13.40 & 11.46 \\
Qwen2.5-7B-VL & 10.07 & 24.39 & 10.28 & 19.43 & -- & 19.26 & 24.31 \\
\midrule
\multicolumn{8}{@{}l}{\textbf{QASPER (full archive), Acc.}}\\*
GPT-5.2 & 55.44 & 76.99 & 75.19 & 75.92 & -- & 71.64 & 74.36 \\
InternVL3-8B & 48.14 & 69.38 & 55.26 & 56.40 & -- & 46.28 & 62.97 \\
Pixtral-12B & -- & -- & -- & -- & -- & -- & -- \\
Qwen2.5-32B-VL & 43.18 & 77.37 & 75.12 & 76.02 & -- & 72.85 & 74.36 \\
Qwen2.5-7B-VL & 35.56 & 68.48 & 65.78 & 58.20 & -- & 64.68 & 65.64 \\
\midrule
\multicolumn{8}{@{}l}{\textbf{QASPER (Table 2 subset), F1}}\\*
GPT-5.2 & 25.56 & 29.64 & 29.61 & 29.19 & -- & 27.86 & 30.15 \\
InternVL3-8B & 17.78 & 16.91 & 13.07 & 15.54 & -- & 11.20 & 16.71 \\
Pixtral-12B & -- & -- & -- & -- & -- & -- & -- \\
Qwen2.5-32B-VL & 12.06 & 12.13 & 10.70 & 12.92 & -- & 10.44 & 11.55 \\
Qwen2.5-7B-VL & 10.60 & 25.44 & 19.45 & 20.73 & -- & 19.14 & 24.26 \\
\midrule
\multicolumn{8}{@{}l}{\textbf{QASPER (Table 2 subset), Acc.}}\\*
GPT-5.2 & 51.92 & 75.10 & 77.25 & 74.27 & -- & 70.88 & 70.47 \\
InternVL3-8B & 44.02 & 69.36 & 54.43 & 56.15 & -- & 44.64 & 62.57 \\
Pixtral-12B & -- & -- & -- & -- & -- & -- & -- \\
Qwen2.5-32B-VL & 37.71 & 76.35 & 76.14 & 76.56 & -- & 71.36 & 70.05 \\
Qwen2.5-7B-VL & 30.49 & 66.32 & 64.38 & 57.05 & -- & 62.09 & -- \\
Qwen3-VL-8B & 53.27 & 76.25 & 70.75 & 56.89 & -- & -- & -- \\
InternVL3.5-8B & 49.38 & 57.66 & 61.07 & 55.68 & -- & -- & -- \\
\midrule
\multicolumn{8}{@{}l}{\textbf{SQuAD v2 (100 matched questions), F1}}\\*
GPT-5.2 & 78.92 & 68.77 & 72.44 & 68.15 & 65.38 & 75.77 & 81.51 \\
InternVL3-8B & 32.59 & 32.17 & 24.98 & 18.63 & 26.12 & 14.83 & 38.04 \\
InternVL3.5-8B & 48.78 & 26.89 & 13.91 & 28.51 & 6.10 & 17.13 & 32.52 \\
Pixtral-12B & 74.17 & 32.72 & 19.24 & 33.07 & 31.58 & 7.06 & 39.02 \\
Qwen2.5-7B-VL & 6.63 & 25.60 & 27.03 & 30.32 & 29.30 & 29.10 & 35.41 \\
Qwen2.5-32B-VL & 13.85 & 25.95 & 20.40 & 26.09 & 24.27 & 28.34 & 29.65 \\
Qwen3-VL-8B & 66.46 & 36.47 & 27.56 & 33.12 & 20.38 & 32.42 & 45.26 \\
\midrule
\multicolumn{8}{@{}l}{\textbf{SQuAD v2 (100 matched questions), accuracy}}\\*
GPT-5.2 & 97.50 & 85.50 & 94.00 & 85.50 & 85.50 & 95.50 & 98.00 \\
InternVL3-8B & 82.50 & 84.00 & 73.50 & 69.50 & 76.50 & 79.50 & 94.00 \\
InternVL3.5-8B & 98.00 & 83.50 & 84.00 & 79.50 & 86.00 & 91.00 & 96.00 \\
Pixtral-12B & 91.50 & 81.00 & 83.50 & 84.50 & 82.50 & 81.00 & 93.00 \\
Qwen2.5-7B-VL & 80.00 & 82.50 & 83.50 & 81.50 & 77.00 & 90.00 & 93.50 \\
Qwen2.5-32B-VL & 86.50 & 84.00 & 87.00 & 83.00 & 83.50 & 95.50 & 95.00 \\
Qwen3-VL-8B & 95.00 & 85.50 & 84.50 & 84.00 & 85.50 & 95.50 & 97.50 \\
\end{longtable}
\endgroup

\section{Evaluation for QASPER and SQuAD v2.}
\label{app:evaluation}
Standard EM or span-matching metrics are insufficient as a primary measure in our setting, since many answers are free-form, paraphrastic, or require multi-sentence justification. We therefore adopt a model-based evaluation for Accuracy: GPT-5 serves as an automatic judge and assigns a score of 1.0 (fully correct), 0.5 (partially correct), or 0.0 (incorrect) by comparing the model prediction against the reference answer. Accuracy is reported as the mean of these scores across all examples.
In addition, we report the F1 used in prior work on QASPER and SQuAD v2. Concretely, answers are normalized (lowercasing, removing punctuation and articles), and F1 is computed as the harmonic mean of precision and recall over overlapping tokens between the predicted and ground-truth answers, taking the maximum over multiple references when available.

\section{Compute Resources}
\label{app:compute}

We report the computational resources used for the main experiments in this paper.

\paragraph{Evaluation experiments.}
All evaluation experiments were conducted on NVIDIA A100/H100 80GB GPUs. Running inference across all 7 models, 7 benchmarks, and 5 input modes required approximately 1000 GPU-hours in total. The majority of compute was consumed by long-context datasets (QASPER and SQuAD), larger models (Qwen2.5-VL-32B and GPT-5.2 via API), and multi-staged inference.
Approximtely 200 GPU-hours were used for the preliminary experiments.

\paragraph{Self-distillation training.}
Each self-distillation run (LoRA fine-tuning with rank 64) was conducted on a single H100 80GB GPUs. Training for 2 epochs on GSM8K ($\sim$7K examples) required approximately 2 hours per model, and training on MMLU ($\sim$7K examples) required approximately 3 hours per model.

\pagebreak
\paragraph{Error analysis.}
The grounded-theory error analysis involved GPT-5.2 API calls for coding 5,245 errors. 
Including preliminary experiments, this consumed approximately \$500 in API costs.

\section{FLOPs Analysis}
\label{app:flops}

To quantify the computational overhead of image-based inputs relative to text, we report the ratio of prefill FLOPs consumed by visual tokens to those consumed by text tokens for the same input.
A higher ratio indicates that the image pathway demands proportionally more compute.
As shown in \Cref{app:tab:flops_ratio}, the ratio varies considerably across models and datasets.
For short-context benchmarks (ARC, GSM8K, HumanEval, MMLU) with a fixed rendered-image resolution, the ratio typically ranges from $1.4\times$ to $4.7\times$, with Pixtral-12B being a notable outlier due to its higher native image-token count.
For long-context datasets (SQuAD and QASPER), where the number of input images scales with document length, the ratio increases substantially, reaching up to $29\times$ for Pixtral-12B on QASPER.
These results highlight that, while visual inputs can provide richer structural signals, they come at a non-trivial computational cost that grows with document length.

\begin{table}[ht]
\centering
\scriptsize
\caption{Image-to-Text FLOPs ratio across models and datasets.
A ratio of $k$ indicates that processing the visual input requires $k\times$ the FLOPs of processing the equivalent text input.
Missing entries indicate that the model does not support the corresponding configuration.}
\label{app:tab:flops_ratio}
\resizebox{\columnwidth}{!}{%
\begin{tabular}{@{}lcccccc@{}}
\toprule
\textbf{Dataset} & \textbf{InternVL3-8B} & \textbf{Pixtral-12B} & \textbf{Qwen2.5-7B-VL} & \textbf{Qwen2.5-32B-VL} & \textbf{Qwen3-VL-8B} & \textbf{Gemma3-4B} \\
\midrule
ARC        & 3.68  & 13.42 & 3.92  & 2.45  & 3.81  & 4.68  \\
GSM8K      & 1.98  & 13.65 & 4.01  & 1.56  & 2.82  & 2.23  \\
HumanEval  & 1.52  & 6.21  & 1.80  & 1.36  & 1.62  & 1.72  \\
MMLU       & 3.45  & 12.78 & --    & 2.28  & 2.74  & 4.11  \\
SQuAD      & 8.64  & 20.70 & 12.37 & 7.54  & 8.98  & 8.98  \\
QASPER     & 8.33  & 29.29 & 9.38  & 5.91  & 6.50  & 1.80  \\
\bottomrule
\end{tabular}%
}
\end{table}

\section{OCR Quality and Task Accuracy}
\label{app:ocr_correlation}

\begin{figure}[t]
  \centering
  \begin{subfigure}{0.48\textwidth}
    \centering
    \includegraphics[width=\linewidth]{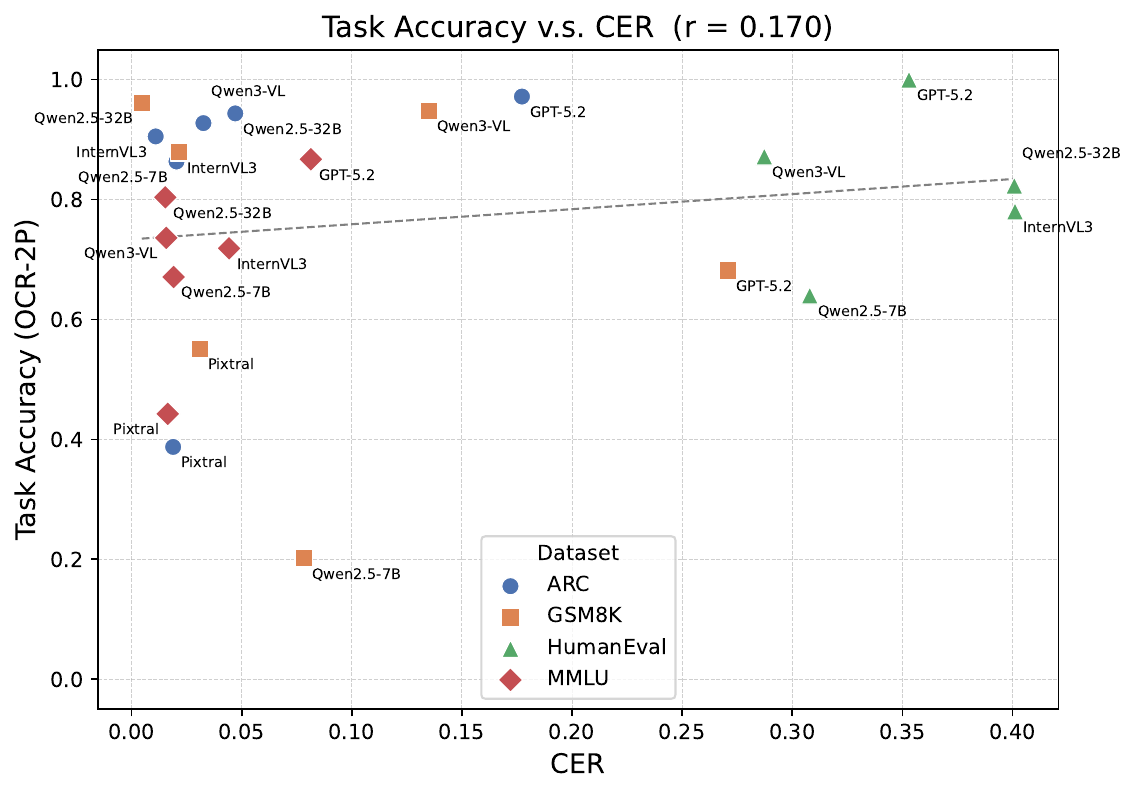}
    \caption{Task accuracy vs.\ CER.}
    \label{fig:accuracy_vs_cer}
  \end{subfigure}
  \hfill
  \begin{subfigure}{0.48\textwidth}
    \centering
    \includegraphics[width=\linewidth]{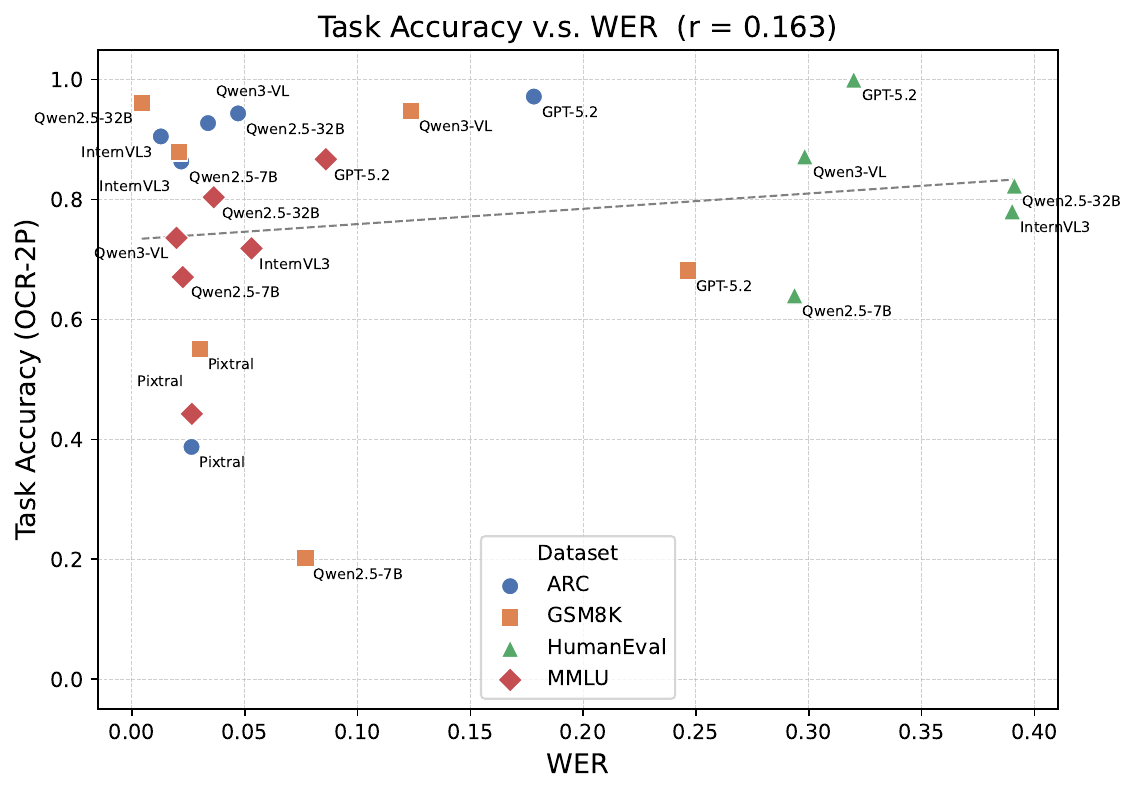}
    \caption{Task accuracy vs.\ WER.}
    \label{fig:accuracy_vs_wer}
  \end{subfigure}
  \caption{Correlation between OCR quality and task accuracy under the \texttt{OCR-2Pass} setting. Each point represents one model--dataset pair. The dashed line shows the linear trend.}
  \label{fig:ocr_correlation}
\end{figure}

To investigate whether the visual recognition capability of each model explains the modality gap, we measure the correlation between OCR quality and task accuracy under the \texttt{OCR-2Pass} setting.
For each model--dataset pair, we compute the Character Error Rate (CER) and Word Error Rate (WER) by comparing the OCR-extracted text against the gold text input, after removing the answer-format instruction sentence and normalizing whitespace.
\Cref{fig:accuracy_vs_cer,fig:accuracy_vs_wer} plot task accuracy against CER and WER, respectively.

The weak correlations ($r = 0.170$ for CER, $r = 0.163$ for WER) indicate that OCR quality alone does not account for the accuracy differences across models and datasets.
The relationship is far from deterministic: several models achieve high accuracy despite moderate OCR errors, and vice versa.
This suggests that the modality gap is driven by factors beyond text recognition, including reasoning collapse and sensitivity to visual formatting.

\subsection{Handwriting Transcription Check}
\label{app:handwriting_ocr}

We conduct a separate transcription-only pass on Priestacy handwriting images for Qwen3-VL-8B and InternVL3-8B on MMLU, GSM8K, and ARC, sampling 50 items per model--dataset pair (300 outputs in total).
We score the transcriptions against the exact text used to render each image and match each item to its handwriting image-mode answer to assess the relationship between transcription quality and task correctness.
Fifteen outputs follow the answer-format instruction embedded in the image instead of the transcription request; these non-transcriptions are excluded from the reported WER/CER summaries but retained in the correlation analysis.

\begin{table}[H]
\centering
\small
\setlength{\tabcolsep}{4pt}
\begin{tabular}{@{}llrrrrr@{}}
\toprule
Model & Dataset & WER (\%) & CER (\%) & \makecell{Image acc.\\(\%)} & $r$ & $p$ \\
\midrule
\multirow{3}{*}{Qwen3-VL-8B}
 & MMLU & 2.7 & 0.6 & 76.0 & $+0.003$ & 0.98 \\
 & GSM8K & 1.8 & 1.4 & 48.0 & $-0.015$ & 0.92 \\
 & ARC & 0.9 & 1.5 & 92.0 & $-0.054$ & 0.71 \\
\midrule
\multirow{3}{*}{InternVL3-8B}
 & MMLU & 6.1 & 1.7 & 42.0 & $-0.215$ & 0.13 \\
 & GSM8K & 1.9 & 0.5 & 52.0 & $-0.144$ & 0.32 \\
 & ARC & 2.2 & 0.8 & 68.0 & $-0.111$ & 0.44 \\
\bottomrule
\end{tabular}
\caption{Handwriting transcription quality and task accuracy on 50 sampled items per model--dataset pair. WER and CER measure the separate transcription-only pass; image accuracy measures answers from the handwriting image-mode evaluation on the matched items. $r$ is the point-biserial correlation between per-item WER and answer correctness, with unadjusted $p$-values. None of the six correlations is significant at $p<0.05$.}
\label{tab:handwriting_ocr}
\end{table}

As shown in Table~\ref{tab:handwriting_ocr}, WER ranges from $0.9\%$ to $6.1\%$ and CER from $0.5\%$ to $1.7\%$, while image-mode accuracy ranges from $42.0\%$ to $92.0\%$ on the matched items.
None of the six point-biserial correlations between per-item WER and answer correctness reaches significance (all unadjusted $p\geq 0.13$).
These results argue against widespread transcription failure as the sole explanation for the handwriting degradation, although this small-sample check cannot exclude effects from errors on individual task-critical tokens.

\section{Inference-Time CoT Prompting in Images}
\label{app:cot_prompting}

As an inference-time intervention for the reluctance to reason under image input, we append ``Please think step-by-step.'' after the question and answer-format instruction in the rendered image.
We evaluate this diagnostic on a subset of our evaluation suite: six open-weight models on all 1{,}319 GSM8K test questions, without additional training.

Since our primary goal is to align image-mode behavior with the model's text-mode reasoning through self-distillation, we report this prompting experiment only in the appendix.

\begin{table}[htbp]
\centering
\small
\setlength{\tabcolsep}{7pt}
\begin{tabular}{@{}lrrrr@{}}
\toprule
\textbf{Model} & \textbf{Pure Text} & \textbf{Pure Image} & \textbf{Image + CoT} & \textbf{Gain (pp)} \\
\midrule
InternVL3-8B     & 78.01 & 42.53 & 70.58 & +28.05 \\
InternVL3.5-8B   & 95.30 & 95.30 & 95.53 & +0.23 \\
Pixtral-12B      & 3.79  & 8.11  & 73.24 & +65.13 \\
Qwen2.5-7B-VL    & 19.48 & 24.79 & 40.41 & +15.62 \\
Qwen2.5-32B-VL   & 95.07 & 94.92 & 95.07 & +0.15 \\
Qwen3-VL-8B      & 93.56 & 30.71 & 86.96 & +56.25 \\
\bottomrule
\end{tabular}
\vspace{2mm}
\caption{GSM8K accuracy (\%) with an explicit CoT instruction rendered inside the image, without additional training ($n=1{,}319$ per model). Pure Text and Pure Image are the original baselines from Table~\ref{tab:overall_across_tasks}; gain is Image + CoT minus Pure Image in percentage points. These are descriptive comparisons; decoding and rendering differences are detailed in Appendix~\ref{app:cot_prompting}.}
\label{tab:cot_prompting}
\end{table}

\mypara{Results.}
Table~\ref{tab:cot_prompting} shows higher accuracy for the CoT-in-image runs than the original \texttt{Pure Image} baselines for all six models.
Qwen3-VL-8B and InternVL3-8B still fall below their text-mode baselines, whereas Qwen2.5-32B-VL and InternVL3.5-8B, whose image-mode baselines are already strong, change by less than 0.3 points.
These results support explicit prompting as a useful inference-time intervention, while motivating training-time alignment that does not depend on an added reasoning cue.

\section{Reasoning Collapse Example}
\label{app:reasoning_collapse_example}

Figure~\ref{fig:reasoning_collapse_example} illustrates the reasoning collapse phenomenon identified in Section~\ref{sec:error_analysis}.
Given the same GSM8K problem, the model produces a detailed step-by-step solution in text mode (593 characters), working through the distance traveled, time constraints, and required speed calculation.
In image mode, the model outputs only an answer tag (19 characters) without any intermediate reasoning, arriving at an incorrect answer.
This example demonstrates how image mode suppresses the reasoning process that would normally catch arithmetic errors and guide the model to the correct solution.

\begin{figure}[h]
\centering
\begin{tcolorbox}[
    enhanced,
    title={\normalsize Reasoning Collapse: Text Mode vs.\ Image Mode},
    colback=gray!3,
    colframe=gray!60,
    left=2mm,right=2mm,
    top=0.6em,bottom=0.6em,
    boxrule=0.4pt,
]
\small
\textbf{Question:} Marissa is hiking a 12-mile trail. She took 1 hour to walk the first 4 miles, then another hour to walk the next two miles. If she wants her average speed to be 4 miles per hour, what speed (in miles per hour) does she need to walk the remaining distance?

\vspace{0.5em}
\hrule
\vspace{0.5em}

\textbf{Text Mode Response} \textcolor{green!60!black}{(\ding{51} Correct)}:

\ttfamily\scriptsize
Marissa has hiked 4 miles in 1 hour, then 2 more miles in another hour. So far, she has hiked 6 miles in 2 hours.

She wants her \textbf{average speed} over the \textbf{entire 12-mile trail} to be \textbf{4 miles per hour}.

Average speed = total distance / total time

So, to average 4 mph over 12 miles, she needs:

Total time = 12 miles / 4 mph = \textbf{3 hours}

She has already used \textbf{2 hours}, so she has \textbf{1 hour left} to hike the remaining \textbf{6 miles}.

Therefore, she needs to walk the remaining 6 miles in 1 hour.

Speed = distance / time = 6 miles / 1 hour = \textbf{6 miles per hour}

<answer>6</answer>

\rmfamily\normalsize
\vspace{0.5em}
\hrule
\vspace{0.5em}

\textbf{Image Mode Response} \textcolor{red!70!black}{(\ding{55} Incorrect)}:

\ttfamily\scriptsize
<answer>12</answer>

\end{tcolorbox}
\caption{Example of reasoning collapse on GSM8K (Qwen3-VL-8B). In text mode, the model produces detailed step-by-step reasoning (593 characters) and arrives at the correct answer. In image mode, the model skips all reasoning and outputs only an incorrect answer (19 characters).}
\label{fig:reasoning_collapse_example}
\end{figure}

\section{Human Annotation Details}
\label{app:human_annotation}

This section describes the human annotators involved in the error analysis (Section~\ref{sec:error_analysis}) and the instructions they received.

\paragraph{Annotator Background.}
The human annotation was conducted by two students who hold graduate degrees in computer science or a related field and have prior experience with qualitative coding methods.
Annotators were fluent in English and familiar with the task domains covered by the benchmarks.

\paragraph{Annotation Procedure.}
The annotation followed a human-in-the-loop grounded theory protocol, as described in Section~\ref{sec:error_analysis}.
The annotators' responsibilities included:
\begin{enumerate}[leftmargin=*,itemsep=2pt]
    \item \textbf{Reviewing proposed codes (Phase 2):} For each new code, merge, or description update proposed by GPT-5.2 during the constant comparison phase, an annotator reviewed the proposal along with the associated error instances. The annotator approved, rejected, or modified the proposal based on whether the code accurately captured the failure mode and was consistent with existing codes.
    \item \textbf{Axial coding (Phase 3):} After saturation was reached, one expert annotator grouped the refined codes into higher-level categories through discussion, producing the final seven-category taxonomy reported in Section~\ref{sec:error_analysis}.
\end{enumerate}

\paragraph{Annotation Guidelines.}
During the open and constant comparison phases, annotators followed these principles:
\begin{itemize}[leftmargin=*,itemsep=2pt]
    \item \textbf{Code granularity:} Codes should be specific enough to capture distinct failure modes but general enough to apply across multiple instances. Avoid overly narrow codes that apply to only one or two errors.
    \item \textbf{Mutual exclusivity:} Each error should be assignable to exactly one code. If an error appears to fit multiple codes, the annotator should either refine the code definitions or create a new code that better captures the failure.
    \item \textbf{Consistency:} When reviewing a proposed code, compare it against existing codes to ensure no conceptual overlap. If overlap exists, consider merging codes or clarifying boundaries.
    \item \textbf{Focus on root cause:} Classify errors based on the primary failure mode, not secondary symptoms. For example, if a calculation error leads to an incorrect final answer, classify it as a calculation error rather than a factual error.
\end{itemize}

\paragraph{Error Classification Prompt.}
Figure~\ref{fig:error-classification-prompt} shows the prompt used to classify errors.
All three annotators (human expert, GPT-5.2, and Claude) used identical guidelines based on this prompt, which defines each error category with explicit criteria and provides decision rules to resolve ambiguous cases.

\begin{figure*}[t]
\centering
\begin{tcolorbox}[
    enhanced,
    title={\normalsize Error Classification Prompt},
    colback=gray!3,
    colframe=gray!60,
    left=2mm,right=2mm,
    fontupper=\small,
    breakable
]
You are an expert error analyst classifying why a model got a question wrong.

\textbf{Categories:}
\begin{enumerate}[leftmargin=*,itemsep=1pt]
\item \textbf{Conceptual/FactualRecall}: The model states or relies on an incorrect fact or misconception that you can explicitly identify. You must be able to say ``The model incorrectly believes X, but the correct fact is Y.'' This includes questions asking about definitions of concepts.

\item \textbf{Reasoning}: The model's logic or inference is flawed. This includes: drawing wrong conclusions, faulty causal reasoning, misapplying a doctrine/rule/principle to a situation, flawed moral/legal/logical judgment, or making analytical errors. Use this when the model reasons through the problem but reaches wrong conclusions without stating an explicit factual error.

\item \textbf{Calculation/Mathematical}: Errors involving numbers or formulas. This includes: arithmetic mistakes, using the wrong formula, miscalculating values, or errors in mathematical reasoning or derivation.

\item \textbf{Incomplete/Partial}: The response is empty, truncated, or lacks a parsable answer.

\item \textbf{Format/Output}: The model's answer content is correct but formatting is wrong (missing tags, wrong structure). Never use this if the answer is wrong.

\item \textbf{Question Interpretation}: The model misread the question itself. Rarely used.

\item \textbf{Incorrect Rationale}: Only use when: (a) the model's final answer exactly matches the gold answer, and (b) the reasoning contradicts the answer. If the answer is wrong, this category cannot apply.

\item \textbf{MISC}: Truly uncategorizable. Rarely used.
\end{enumerate}

\textbf{Critical Rules:}
\begin{itemize}[leftmargin=*,itemsep=1pt]
\item First check if the model's answer matches the gold answer. If the answer is wrong, you cannot use ``Incorrect Rationale'' or ``Format/Output''.
\item Most errors are either Conceptual/FactualRecall or Reasoning. These two categories cover $\sim$90\% of errors.
\item Conceptual/FactualRecall vs Reasoning distinction: Conceptual/FactualRecall = model states an incorrect fact; Reasoning = model misapplies correct knowledge.
\item Calculation/Mathematical: Use for any error involving numbers, formulas, or mathematical reasoning.
\end{itemize}

First, verify: Does the model's answer match the gold answer? Then classify the error type.
\end{tcolorbox}
\caption{Prompt used for error classification in the inter-annotator agreement study.}
\label{fig:error-classification-prompt}
\end{figure*}

\paragraph{Inter-Annotator Agreement.}
To validate the reliability of our error taxonomy, we conducted an inter-annotator agreement study with three annotators: the agreement of two human experts, GPT-5.2, and Claude (claude-opus-4-5-2025-11-01).
We sampled 50 errors with human labels and had each annotator independently classify them using the prompt in Figure~\ref{fig:error-classification-prompt}.
Table~\ref{tab:inter-annotator} reports pairwise Cohen's kappa ($\kappa$) and raw agreement percentages.

\begin{table}[h]
\centering
\small
\begin{tabular}{lcc}
\toprule
\textbf{Annotator Pair} & \textbf{Cohen's $\kappa$} & \textbf{Raw Agreement} \\
\midrule
Human vs.\ GPT-5.2 & 0.57 (Moderate) & 75.0\% \\
Human vs.\ Claude & 0.69 (Substantial) & 81.8\% \\
GPT-5.2 vs.\ Claude & 0.64 (Substantial) & 79.5\% \\
\midrule
Three-way agreement & --- & 68.2\% \\
\bottomrule
\end{tabular}
\caption{Inter-annotator agreement for error classification. Cohen's $\kappa$ values indicate moderate to substantial agreement across all annotator pairs.}
\label{tab:inter-annotator}
\end{table}

The substantial agreement between human and model annotators ($\kappa \geq 0.57$) suggests that our error taxonomy is well-defined and can be reliably applied.
The three-way agreement of 68.2\% indicates that the majority of errors receive consistent classifications across all annotators, with remaining disagreements occurring at inherently ambiguous boundaries (e.g., distinguishing conceptual errors from reasoning errors when a model misapplies domain knowledge).

\section{Limitations}
\label{app:limitations}

We acknowledge several limitations of this work.

\paragraph{Causality of the reasoning collapse}
Our diagnostic claim is that image-mode performance degradation is associated primarily with shortened reasoning chains rather than with perceptual or knowledge failures. 
We support this through three converging lines of evidence: output-length statistics, the error-category distribution shift (Figure 6a), and the recovery achieved by restoring long reasoning traces via self-distillation. 
While this kind of approach is common in empirical research, we would like to emphasize that this evidence is \textit{correlational and intervention-based} rather than fully causal. 
Several additional plausible mechanisms could co-explain the observed pattern, and our experiments do not cleanly separate them. 
Disentangling distributional, perceptual, and behavioral contributions is an important direction we leave to future work.

\paragraph{Rapidly evolving model landscape.}
The MLLM field is advancing at an unprecedented pace, with model providers continually refining their training recipes, data curation strategies, and architectural designs.
Our evaluation captures a snapshot of seven models at a particular point in time.
Indeed, we observe that newer models such as InternVL3.5-8B already exhibit smaller modality gaps, suggesting that some of the issues we identify may be addressed through improved training procedures.
However, the underlying question of \emph{how} multi-stage training paradigms and language processing methods, which differ fundamentally from human perceptual learning, shape a model's visual text understanding remains open.
Our diagnostic framework and error taxonomy provide tools for continued investigation as the field evolves.

\paragraph{Scope of models and benchmarks.}
While we evaluate seven MLLMs spanning different architectures and scales, this represents a fraction of the available models.
Our findings may not generalize to architectures with fundamentally different vision-language fusion mechanisms (e.g., early fusion, pixel-based tokenization).
Similarly, our benchmark selection focuses on reasoning-intensive tasks; performance patterns on other task types (e.g., visual grounding, multi-image reasoning) may differ.

\paragraph{Statistical reporting.}
Our HumanEval results use single-sample pass@1 estimates over only 164 problems, which is well known to be high-variance.
While we discuss this caveat, we did not generate multiple samples per problem and report unbiased pass@k estimates. 
HumanEval results should therefore be read as illustrative of qualitative trends rather than as precise capability measurements. 
For our larger benchmarks (MMLU, ARC, GSM8K, GPQA), the sample sizes support more confident conclusions, but we did not run multiple seeds for the self-distillation experiments due to compute constraints and report results from single training runs.

\paragraph{Zero-shot evaluation only.}
All experiments are conducted in a zero-shot setting to avoid confounding factors from in-context examples.
However, few-shot prompting or chain-of-thought demonstrations may mitigate the reasoning collapse we observe while introducing additional confounding factors.
Future work should investigate whether the modality gap persists under different prompting strategies.

\paragraph{Grounded-theory methodology.}
Our error analysis follows Nelson's computational grounded theory framework, with GPT-5.2 performing initial open coding and constant comparison and human annotators reviewing every proposed code change and conducting axial coding. 
While adopting the approach from prior work, we acknowledge that this allocation differs slightly from classical grounded theory, where human interpretation drives the most generative coding phase. 
Our human–GPT-5.2 agreement of $\kappa= 0.57$ is "moderate" rather than "substantial," and the boundary between Conceptual/Factual and Reasoning errors—precisely the distinction underlying our central claim that image mode preserves knowledge and reasoning capability—is the most contested. 
Human–Claude agreement is higher ( $\kappa = 0.69$), and three-way agreement is 68.2\%, which we view as adequate for taxonomy validation but not fully eliminating interpretive ambiguity. 
Additionally, with ~150 errors per model–dataset cell, fine-grained per-cell claims about error distributions are statistically limited; our findings are most reliable in aggregate and should be interpreted in aggregate.

\clearpage
\section{Output Lengths by Correctness}
\label{app:output_lengths}

Tables~\ref{tab:lengths_distilled} and~\ref{tab:length_ratios} compare response lengths before and after self-distillation, measured in whitespace-delimited words and split by answer correctness.
Self-distillation often restores comparable text/image lengths: for Qwen3-VL-8B on GSM8K, the ratio falls from $33.04$ to $1.02$ after GSM8K training, though recovery is not universal.
InternVL3-8B on MMLU often transcribes the image before answering (122.3 words in image mode versus 4.8 in text mode), so length alone does not measure reasoning quality.

\begingroup
\fontsize{8.5}{10}\selectfont
\setlength{\tabcolsep}{4pt}
\renewcommand{\arraystretch}{1.12}
\begin{longtable}{@{}llrrr@{\hspace{14pt}}rrr@{}}
\caption{Mean response lengths (words) before and after self-distillation, split by answer correctness. T/I denote Pure Text/Pure Image; S/V denote MMMU-Pro Standard/Vision. A dash indicates unavailable outputs. Correct and incorrect means are conditioned separately within each stage.}\label{tab:lengths_distilled}\\
\toprule
& & \multicolumn{3}{c}{\textbf{Baseline}} & \multicolumn{3}{c}{\textbf{Self-distilled}} \\
\cmidrule(lr){3-5}\cmidrule(lr){6-8}
Benchmark & Mode & All & Correct & Incorrect & All & Correct & Incorrect \\
\midrule
\endfirsthead
\multicolumn{8}{l}{\tablename\ \thetable\ (continued)}\\
\toprule
& & \multicolumn{3}{c}{\textbf{Baseline}} & \multicolumn{3}{c}{\textbf{Self-distilled}} \\
\cmidrule(lr){3-5}\cmidrule(lr){6-8}
Benchmark & Mode & All & Correct & Incorrect & All & Correct & Incorrect \\
\midrule
\endhead
\midrule
\multicolumn{8}{r}{Continued on next page}\\
\endfoot
\bottomrule
\endlastfoot
\rowcolor{black!6}
\multicolumn{8}{@{}l}{\strut\textbf{Qwen3-VL-8B}\quad Distilled on GSM8K}\\*
GSM8K & T & 121.5 & 93.9 & 521.4 & 112.3 & 86.0 & 530.3 \\*
 & I & 3.7 & 6.7 & 2.4 & 109.9 & 85.3 & 423.3 \\
\addlinespace[2pt]
ARC & T & 8.7 & 9.0 & 5.3 & 2.0 & 1.8 & 4.1 \\*
 & I & 3.5 & 3.5 & 3.2 & 8.4 & 8.0 & 11.9 \\
\addlinespace[2pt]
MMLU & T & 38.7 & 35.5 & 50.3 & 7.9 & 6.4 & 12.3 \\*
 & I & 7.7 & 7.8 & 7.4 & 40.2 & 27.1 & 75.0 \\
\addlinespace[2pt]
MMMU-Pro & S & 1512.4 & 932.3 & 2399.5 & 1437.9 & 790.0 & 2353.3 \\*
 & V & 1580.7 & 889.7 & 2149.3 & 1413.9 & 692.1 & 1992.8 \\
\addlinespace[5pt]
\rowcolor{black!6}
\multicolumn{8}{@{}l}{\strut\textbf{Qwen3-VL-8B}\quad Distilled on MMLU}\\*
GSM8K & T & 121.5 & 93.9 & 521.4 & 136.6 & 118.5 & 402.4 \\*
 & I & 3.7 & 6.7 & 2.4 & 136.9 & 118.8 & 239.6 \\
\addlinespace[2pt]
ARC & T & 8.7 & 9.0 & 5.3 & 216.4 & 210.0 & 282.8 \\*
 & I & 3.5 & 3.5 & 3.2 & 211.0 & 203.9 & 288.1 \\
\addlinespace[2pt]
MMLU & T & 38.7 & 35.5 & 50.3 & 262.1 & 234.0 & 335.0 \\*
 & I & 7.7 & 7.8 & 7.4 & 258.1 & 228.6 & 333.2 \\
\addlinespace[2pt]
MMMU-Pro & S & 1512.4 & 932.3 & 2399.5 & 1702.0 & 978.3 & 2700.5 \\*
 & V & 1580.7 & 889.7 & 2149.3 & 1907.8 & 979.0 & 2633.8 \\
\addlinespace[5pt]
\rowcolor{black!6}
\multicolumn{8}{@{}l}{\strut\textbf{InternVL3-8B}\quad Distilled on GSM8K}\\*
GSM8K & T & 131.4 & 141.5 & 95.6 & 148.5 & 146.0 & 171.7 \\*
 & I & 142.0 & 144.4 & 140.2 & 147.3 & 145.7 & 162.0 \\
\addlinespace[2pt]
ARC & T & 13.1 & 12.6 & 18.8 & 97.4 & 94.4 & 129.9 \\*
 & I & 2.4 & 2.3 & 3.3 & 41.2 & 38.3 & 65.6 \\
\addlinespace[2pt]
MMLU & T & 4.8 & 4.3 & 5.5 & 119.5 & 105.9 & 160.7 \\*
 & I & 122.3 & 111.0 & 128.1 & 77.7 & 62.1 & 111.0 \\
\addlinespace[2pt]
MMMU-Pro & S & 215.2 & 156.3 & 269.7 & 270.3 & 203.6 & 350.3 \\*
 & V & 194.7 & 150.9 & 216.6 & 300.2 & 213.9 & 341.9 \\
\addlinespace[5pt]
\rowcolor{black!6}
\multicolumn{8}{@{}l}{\strut\textbf{InternVL3-8B}\quad Distilled on MMLU}\\*
GSM8K & T & 131.4 & 141.5 & 95.6 & -- & -- & -- \\*
 & I & 142.0 & 144.4 & 140.2 & 149.7 & 145.4 & 175.2 \\
\addlinespace[2pt]
ARC & T & 13.1 & 12.6 & 18.8 & 182.9 & 181.1 & 201.3 \\*
 & I & 2.4 & 2.3 & 3.3 & 174.5 & 172.9 & 190.3 \\
\addlinespace[2pt]
MMLU & T & 4.8 & 4.3 & 5.5 & 217.4 & 205.0 & 251.9 \\*
 & I & 122.3 & 111.0 & 128.1 & 201.1 & 187.8 & 236.1 \\
\addlinespace[2pt]
MMMU-Pro & S & 215.2 & 156.3 & 269.7 & 301.0 & 243.3 & 372.3 \\*
 & V & 194.7 & 150.9 & 216.6 & 334.7 & 263.3 & 372.6 \\
\addlinespace[5pt]
\end{longtable}
\endgroup

\begin{table}[htbp]
\centering
\small
\setlength{\tabcolsep}{5pt}
\renewcommand{\arraystretch}{1.2}
\caption{Text/image mean-length ratios before and after self-distillation. GSM8K and MMLU identify the distillation data; 1.00 denotes equal mean lengths. $^{\star}$MMMU-Pro uses Standard/Vision ratios as a descriptive forgetting control. Ratios use unrounded means; a dash indicates unavailable outputs.}\label{tab:length_ratios}
\begin{tabular}{@{}lrrr@{\hspace{16pt}}rrr@{}}
\toprule
& \multicolumn{3}{c}{\textbf{Qwen3-VL-8B}} & \multicolumn{3}{c}{\textbf{InternVL3-8B}} \\
\cmidrule(lr){2-4}\cmidrule(lr){5-7}
& & \multicolumn{2}{c}{Self-distilled} & & \multicolumn{2}{c}{Self-distilled} \\
\cmidrule(lr){3-4}\cmidrule(lr){6-7}
Benchmark & Baseline & GSM8K & MMLU & Baseline & GSM8K & MMLU \\
\midrule
GSM8K & 33.04 & 1.02 & 1.00 & 0.93 & 1.01 & -- \\
ARC & 2.48 & 0.24 & 1.03 & 5.55 & 2.37 & 1.05 \\
MMLU & 5.04 & 0.20 & 1.02 & 0.04 & 1.54 & 1.08 \\
\midrule
MMMU-Pro$^{\star}$ & 0.96 & 1.02 & 0.89 & 1.11 & 0.90 & 0.90 \\
\bottomrule
\end{tabular}
\vspace{8pt}
\end{table}

\section{Significance Tests}
\label{app:significance}

We mark Holm-adjusted significance ($p<0.05$) directly in the result tables.
Tests pair the same benchmark items: two-sided exact McNemar for binary correctness, and 10{,}000 paired bootstrap resamples for gap closure and fractional HumanEval rendering scores.
Gap closure is $(A_{\mathrm{T}}-A_{\mathrm{I}})_{\mathrm{base}}-(A_{\mathrm{T}}-A_{\mathrm{I}})_{\mathrm{distilled}}$, resampling all four outcomes together.
Holm families comprise 35 text--image and 105 other text--mode contrasts; 12 tests each for baseline gaps, residual gaps, and signed gap closure; and four tests per MMMU-Pro configuration.
Unavailable tests retain their place in the planned families (set to $p=1$ for adjustment only).
Rendering tests use a separate 60-test family (Appendix~\ref{app:rendering_significance}).

For Figure~\ref{fig:error_analysis}, we compare image/text error shares within model--benchmark pairs having at least ten coded errors in each mode, using 10{,}000 within-pair label permutations and Holm correction across eight categories.
Format, incomplete/partial, and miscellaneous error shares differ significantly; the calculation-error contrast does not survive this benchmark-matched test.
Estimated absolute category-error rates weight each pair's error share by its full-evaluation error rate.

\clearpage

\newpage
\section*{NeurIPS Paper Checklist}

\begin{enumerate}

\item {\bf Claims}
    \item[] Question: Do the main claims made in the abstract and introduction accurately reflect the paper's contributions and scope?
    \item[] Answer: \answerYes{}
    \item[] Justification: The abstract and introduction claim that we (1) systematically diagnose the modality gap across 7 models, 7 benchmarks, and 5 input modes (Section~3--4), (2) identify reasoning collapse as the root cause through error analysis of 4,000+ examples (Section~5), and (3) propose self-distillation to close the gap (Section~6). All claims are supported by experimental results in the corresponding sections.

\item {\bf Limitations}
    \item[] Question: Does the paper discuss the limitations of the work performed by the authors?
    \item[] Answer: \answerYes{}
    \item[] Justification: We include a dedicated Limitations section in Appendix~\ref{app:limitations}, discussing: (1) the rapidly evolving model landscape and how multi-stage training paradigms may affect visual text understanding, (2) scope of models and benchmarks evaluated, (3) zero-shot evaluation only, (4) self-distillation tested on two architectures, and (5) annotation completeness.

\item {\bf Theory assumptions and proofs}
    \item[] Question: For each theoretical result, does the paper provide the full set of assumptions and a complete (and correct) proof?
    \item[] Answer: \answerNA{}
    \item[] Justification: This is an empirical paper that does not include theoretical results, theorems, or formal proofs. All contributions are based on experimental evaluation and qualitative analysis.

    \item {\bf Experimental result reproducibility}
    \item[] Question: Does the paper fully disclose all the information needed to reproduce the main experimental results of the paper to the extent that it affects the main claims and/or conclusions of the paper (regardless of whether the code and data are provided or not)?
    \item[] Answer: \answerYes{}
    \item[] Justification: We provide: model names and versions (Section~3.3), dataset descriptions and sources (Section~3.1), image rendering specifications including resolution 1280$\times$720 and fonts (Section~3.2), decoding parameters in Appendix (temperature, batch size, max generation length), LoRA hyperparameters for self-distillation (Section~6), and detailed input format examples for all datasets (Appendix~A--B). We also release all the code and data to reproduce all results.

\item {\bf Open access to data and code}
    \item[] Question: Does the paper provide open access to the data and code, with sufficient instructions to faithfully reproduce the main experimental results, as described in supplemental material?
    \item[] Answer: \answerYes{}
    \item[] Justification: We release code and data at \url{https://github.com/KaiserWhoLearns/ReadingNotThinking}. All datasets used (MMLU, ARC, GPQA, GSM8K, HumanEval, QASPER, SQuAD) are publicly available. The repository includes evaluation scripts, image rendering code, and self-distillation training code with instructions to reproduce all experimental results.

\item {\bf Experimental setting/details}
    \item[] Question: Does the paper specify all the training and test details (e.g., data splits, hyperparameters, how they were chosen, type of optimizer) necessary to understand the results?
    \item[] Answer: \answerYes{}
    \item[] Justification: We specify decoding parameters in the Appendix, LoRA hyperparameters for self-distillation (rank $r{=}64$, learning rate $2{\times}10^{-4}$, 2 epochs, batch size 16), evaluation metrics for each dataset in the experimental setup section, and input format specifications in Appendix.

\item {\bf Experiment statistical significance}
    \item[] Question: Does the paper report error bars suitably and correctly defined or other appropriate information about the statistical significance of the experiments?
    \item[] Answer: \answerYes{}
    \item[] Justification: We use bootstrap significance testing for our evaluation results to assess statistical significance of the observed performance differences across modalities and models.

\item {\bf Experiments compute resources}
    \item[] Question: For each experiment, does the paper provide sufficient information on the computer resources (type of compute workers, memory, time of execution) needed to reproduce the experiments?
    \item[] Answer: \answerYes{}
    \item[] Justification: We report GPU types, GPU-hours, and API costs in Appendix~\ref{app:compute}. FLOPs analysis comparing modalities is provided in Appendix~\ref{app:flops}.
    
\item {\bf Code of ethics}
    \item[] Question: Does the research conducted in the paper conform, in every respect, with the NeurIPS Code of Ethics \url{https://neurips.cc/public/EthicsGuidelines}?
    \item[] Answer: \answerYes{}
    \item[] Justification: We have reviewed the NeurIPS Code of Ethics. Our research evaluates publicly available models on public benchmarks, proposes a training method to improve model capabilities, and does not involve human subjects, deception, or potential for direct harm.

\item {\bf Broader impacts}
    \item[] Question: Does the paper discuss both potential positive societal impacts and negative societal impacts of the work performed?
    \item[] Answer: \answerNA{}
    \item[] Justification: This is foundational research diagnosing and improving visual text understanding in multimodal LLMs. The work does not introduce new capabilities with direct negative societal implications beyond those already present in the underlying models. Improvements in visual text understanding are broadly beneficial for accessibility and document processing.
    
\item {\bf Safeguards}
    \item[] Question: Does the paper describe safeguards that have been put in place for responsible release of data or models that have a high risk for misuse (e.g., pre-trained language models, image generators, or scraped datasets)?
    \item[] Answer: \answerNA{}
    \item[] Justification: The paper does not release pre-trained models with high misuse risk. We release evaluation code and LoRA adapters for self-distillation, which are lightweight modifications to existing public models and do not introduce new misuse vectors.

\item {\bf Licenses for existing assets}
    \item[] Question: Are the creators or original owners of assets (e.g., code, data, models), used in the paper, properly credited and are the license and terms of use explicitly mentioned and properly respected?
    \item[] Answer: \answerYes{}
    \item[] Justification: All datasets (MMLU, ARC, GPQA, GSM8K, HumanEval, QASPER, SQuAD) and models (Qwen-VL, InternVL, Pixtral, GPT-5.2) are properly cited with their original papers. All datasets used are publicly available for research purposes.

\item {\bf New assets}
    \item[] Question: Are new assets introduced in the paper well documented and is the documentation provided alongside the assets?
    \item[] Answer: \answerYes{}
    \item[] Justification: We release evaluation code, image rendering scripts, and self-distillation training code as supplementary material. Documentation includes instructions for reproducing experiments and details about data preparation.

\item {\bf Crowdsourcing and research with human subjects}
    \item[] Question: For crowdsourcing experiments and research with human subjects, does the paper include the full text of instructions given to participants and screenshots, if applicable, as well as details about compensation (if any)?
    \item[] Answer: \answerNA{}
    \item[] Justification: The paper does not involve crowdsourcing. Human annotation for error analysis was conducted by domain experts (described in Appendix~\ref{app:human_annotation}).

\item {\bf Institutional review board (IRB) approvals or equivalent for research with human subjects}
    \item[] Question: Does the paper describe potential risks incurred by study participants, whether such risks were disclosed to the subjects, and whether Institutional Review Board (IRB) approvals (or an equivalent approval/review based on the requirements of your country or institution) were obtained?
    \item[] Answer: \answerNA{}
    \item[] Justification: The paper does not involve research with human subjects. The human annotation was performed by paper authors analyzing model outputs, which does not constitute human subjects research.

\item {\bf Declaration of LLM usage}
    \item[] Question: Does the paper describe the usage of LLMs if it is an important, original, or non-standard component of the core methods in this research? Note that if the LLM is used only for writing, editing, or formatting purposes and does \emph{not} impact the core methodology, scientific rigor, or originality of the research, declaration is not required.
    \item[] Answer: \answerYes{}
    \item[] Justification: We use LLMs as part of our core methodology: (1) GPT-5 as an LLM-as-judge for evaluating QASPER and SQuAD (Section~3.1), and (2) GPT-5.2 for part of the grounded-theory error analysis coding process (Section~5.1). Both usages are explicitly described in the paper.

\end{enumerate}

\end{document}